\documentclass{article}

\usepackage[preprint]{corl_2026} % Uncomment for pre-prints (e.g., arxiv); This is like ``final'', but will remove the CORL footnote.

\usepackage{amsmath}
\usepackage{amssymb}
\usepackage{booktabs} 
\usepackage{multirow}
\usepackage{multicol}
\usepackage{xspace}
\usepackage{graphicx}
\usepackage{subfiles}
\usepackage{xcolor}
\usepackage{algorithm}
\usepackage{algpseudocode}
\usepackage{subcaption}
\usepackage{caption}
\usepackage{pgfplots}
\usepackage{wrapfig}
\usepackage{tcolorbox}
\tcbuselibrary{skins}
\usepgfplotslibrary{groupplots}
\pgfplotsset{compat=1.18}

\usepackage[table]{xcolor}
\definecolor{topcolor}{RGB}{255, 0, 0} % define as red, change as needed

\definecolor{lightgray}{gray}{0.92}
\definecolor{topcolor}{HTML}{A68D58}
\definecolor{botcolor}{HTML}{476D3D}

\title{World Action Models Enable Continual Imitation Learning with Recurrent Generative Replays}
\author{
  Manish Kumar Govind, Dominick Reilly, Smit Patel, Hieu Le, Srijan Das\\
  Department of Computer Science\\
  University of North Carolina at Charlotte, United States\\
  \texttt{\{mgovind, sdas24\}@charlotte.edu} \\
\url{https://manishgovind.github.io/REGEN/}
}

\newcommand{\methodname}{Recurrent Generative Replay\xspace}
\newcommand{\methodabbrev}{\textsc{ReGen}\xspace}
\begin{document}
\maketitle

%===============================================================================
% \begin{figure}[h]
% \centering
% \scalebox{0.75}{
% \includegraphics[width=\columnwidth]{Figures/REGEN-Teaser.pdf}
% }
% \caption{Teaser}
% \label{fig:teaser}
% \end{figure}

\begin{abstract}
    %World Action Models (WAMs) extend robot policies beyond action prediction by jointly modeling future visual observations of the environment. We present \methodname{} (\methodabbrev), a replay-based continual imitation learning framework that leverages this generative capability to eliminate the need for storing human demonstrations from previous tasks. 
    Going beyond predicting robot actions, World Action Models (WAMs) can also generate future visual observations. We build on this generative capability to propose \methodname (\methodabbrev), a continual imitation learning framework that synthesizes pseudo-replay trajectories, enabling a robot policy to rehearse previously learned tasks without storing their original human demonstrations.
    During continual adaptation, \methodabbrev recursively queries the WAM to synthesize pseudo-replay trajectories conditioned only on prior task instructions and current-task observations.
    Experiments in both simulation and real-world manipulation settings show that \methodabbrev reduces catastrophic forgetting by up to $50\%$ relative to sequential fine-tuning, while approaching the performance of privileged experience replay methods that require access to real replay data. Finally, we analyze the factors limiting generated replay, identifying long-horizon visual degradation and action-observation inconsistency as the primary bottlenecks. Our results establish WAMs as a promising foundation for continual robot learning without stored demonstrations.

    % , constructing pseudo-replays of previous tasks that do not require human demonstration or intervention. 
    % Instead of retaining human demonstrations of previous tasks
\end{abstract}

\keywords{Continual Imitation Learning, World Action Model, Generative Replay, Robot Control} 

\begin{center}
\small\itshape
``A robot that can imagine its past can continue learning its future.''
\end{center}

%===============================================================================

\section{Introduction}

Recently, World Action Models (WAMs) have emerged as a promising paradigm for robot imitation learning by unifying \emph{perception}, \emph{prediction}, and \emph{control} within a single generative framework~\cite{ye2026dreamzero,cosmos-policy,lingbot-va2026,ye2026gigaworld}. 
%As the robotics community increasingly adopts WAMs as generalist robot policies and adapts them to new tasks, continual imitation learning becomes a natural and pressing challenge: unsurprisingly, naively fine-tuning WAMs on new tasks leads to catastrophic forgetting~\cite{french1999catastrophic,luo2025catastrophic_forgetting_study,shenfeld2026rls_razor} of previously learned ones.
% an important question arises: \emph{do WAMs catastrophically forget previously learned behaviors during continual adaptation?} We find that the answer is yes. 
% Continual imitation learning remains a fundamental challenge in robot policy learning, since robots deployed in real-world environments must continually acquire new skills while retaining previously learned behaviors~\cite{thrun1995lifelong,li2016learningwithoutforgetting,liu2023libero}. Motivated by this challenge, this work investigates how to mitigate catastrophic forgetting in WAMs~\cite{french1999catastrophic,luo2025catastrophic_forgetting_study,shenfeld2026rls_razor}.
%In this work, we ask whether the unique generative properties of WAMs can themselves be exploited to mitigate this catastrophic forgetting.
As WAMs become general-purpose robot policies, continual adaptation to new tasks becomes inevitable. Yet, like other learned policies, they suffer from catastrophic forgetting when fine-tuned sequentially~\cite{french1999catastrophic,luo2025catastrophic_forgetting_study,shenfeld2026rls_razor}. This raises a key question: \textit{can the same generative capabilities that make WAMs powerful also serve as a mechanism for retaining previously learned skills?}

Existing approaches for mitigating catastrophic forgetting span two generations of robot policies: conventional visuomotor policies~\cite{chi2024diffusionpolicyvisuomotorpolicy,zhao2023action_chunking_transformer} and more recent Vision-Language-Action (VLA) models~\cite{kim_openvla_2024,black_pizero_2024,pi05}. Across both paradigms, experience replay has emerged as the dominant continual learning strategy, where demonstrations from previous tasks are stored and replayed during adaptation to new tasks~\cite{liu2023libero,zhu2022bottom,wan2024lotuscontinualimitationlearning,liu2026longlivedrobotscontinuallearning,stellarvla2026}. While highly effective, these approaches fundamentally rely on access to ground-truth demonstrations from prior tasks.
This assumption is becoming increasingly impractical in modern robot learning. Earlier progress in robotics was driven by open datasets of robot demonstrations~\cite{open_x_embodiment_rt_x_2023,bu2025agibot_world_dataset,tian2025interndataa1}, whereas the emerging paradigm is centered around pretrained robot foundation models trained on large-scale proprietary datasets that are rarely released publicly~\cite{black_pizero_2024,pi05,ye2026dreamzero}. Consequently, what we desire for continual learning is a replay mechanism that does not depend on access to \emph{true} demonstrations from previous tasks.
% What continual learning therefore needs is a replay mechanism that does not depend on stored demonstrations at all, one where the model itself serves as the source of prior-task data

\begin{wrapfigure}{r}{0.5\textwidth}
    \centering
    \includegraphics[width=\linewidth]{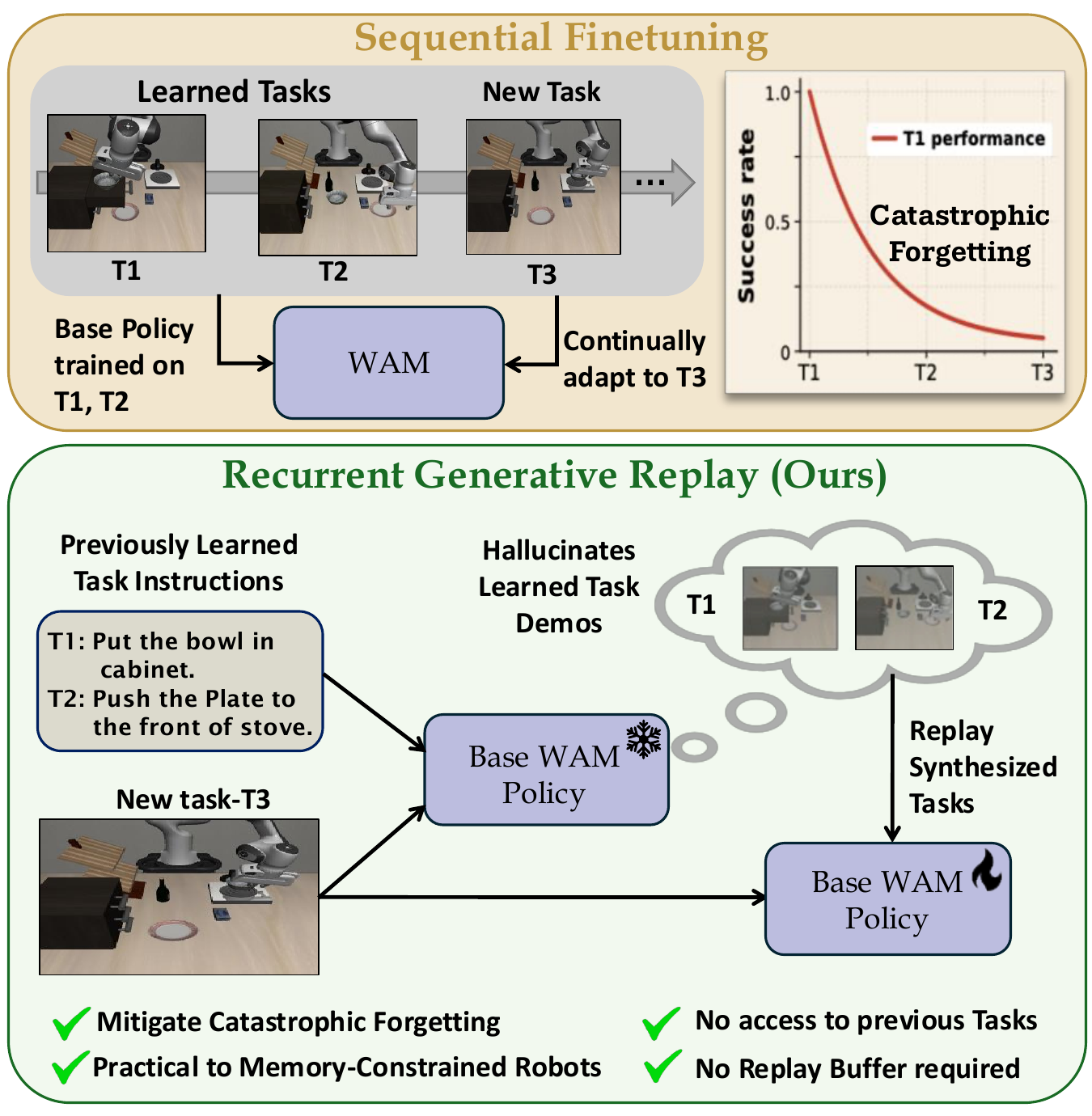}
    \caption{\textbf{Overview of \methodabbrev.} Sequential fine-tuning of WAMs leads to catastrophic forgetting \textcolor{topcolor}{(top)}. \methodabbrev leverages the WAM's generative capabilities to hallucinate pseudo-demonstrations of previously learned tasks, replaying them alongside new task data to mitigate forgetting without storing any prior-task demonstrations. \textcolor{botcolor}{(bottom)}.}
    \label{fig:teaser}
\end{wrapfigure}

WAMs are uniquely positioned to address exactly this challenge. By jointly modeling actions together with future visual observations of the scene~\cite{ye2026dreamzero,cosmos-policy,lingbot-va2026,ye2026gigaworld}, WAMs expose a generative interface capable of synthesizing trajectories for previously learned tasks conditioned only on their language instructions and current observations. In this work, we leverage this property to enable continual imitation learning \emph{without} access to any real demonstrations from previous tasks as shown in Figure~\ref{fig:teaser}.

% without leveraging any previous task demonstrations.
%  \vspace{-0.1in}

Consequently, we propose \textbf{Re}current \textbf{Gen}erative Replay (\textbf{\methodabbrev}), the first continual learning framework that leverages the WAM itself as a native generative replay mechanism.
When adapting to a new task, \methodabbrev generates pseudo-demonstrations for previous tasks by conditioning the WAM on prior task instructions and initial visual observations, then recurrently feeding back its own generated future observations.
%\methodabbrev generates pseudo-demonstrations for previous tasks by conditioning the WAM on prior task instructions, initial visual observations and recurrently feeding back its own generated future observations, 
This yields completed synthetic trajectories of previous tasks. These pseudo-demonstrations are then combined with the new-task demonstrations during fine-tuning to substantially mitigate catastrophic forgetting without requiring human-collected demonstrations of past tasks. 
We evaluate \methodabbrev in both simulated~\cite{liu2023libero} and real-world environments, where it reduces forgetting by over $50\%$ relative to naive sequential fine-tuning and approaches the performance of experience replay methods that rely on privileged access to human-collected demonstrations from previous tasks.
Critically, our analysis reveals that the effectiveness of \methodabbrev is fundamentally tied to two key limitations of current WAMs: \textbf{(i)} degradation in the visual fidelity of future observations during long-horizon recurrent generation, and \textbf{(ii)} inconsistencies between predicted future observations and the corresponding generated actions. These findings indicate that, while \methodabbrev is not yet a complete substitute for real replay, the remaining performance gap is primarily governed by the generative limitations of WAMs themselves. Consequently, improving the fidelity and consistency of WAM generation represents a key direction for closing the gap between pseudo-replay and real experience replay.

We summarize our contributions as follows:
\begin{itemize}
    \item We propose \textbf{\methodname}, the first replay-based continual imitation learning framework that leverages the generative capabilities of WAMs to synthesize replay trajectories without storing real demonstrations from previous tasks.
    
    \item We demonstrate that \methodabbrev substantially mitigates catastrophic forgetting in both simulated and real-world manipulation settings while preserving strong forward transfer.

    \item We identify the primary limitations preventing generated replay from matching real experience replay, namely long-horizon degradation in visual generation and inconsistencies between predicted observations and actions, highlighting key future directions for WAMs.
\end{itemize}

%% Recently... [bring up WAMs, end with bold we exploit this unique property]

%% Our method and contributions

%===============================================================================
\section{Related Work}
\label{sec:related_work}

\paragraph{Continual Imitation Learning.} 
Continual learning aims to acquire a sequence of tasks without catastrophically forgetting previously learned skills~\citep{french1999catastrophic}. Existing approaches fall into three categories: \emph{regularization-based} methods that constrain weight updates~\citep{kirkpatrick2017overcoming, synaptic-int}, \emph{rehearsal-based} methods that retain real samples from prior tasks~\citep{ER}, and \emph{architecture-based} methods that allocate task-specific parameters~\citep{mallya2018packnetaddingmultipletasks, rusu2016progressive}. Prior work~\citep{liu2023libero} adapts these to lifelong robot manipulation, and recent methods extend them with skill discovery, knowledge-distillation, or task-specific adapters combined with replay~\citep{wan2024lotuscontinualimitationlearning, m2distill, liu2024tailtaskspecificadaptersimitation, clare}; a parallel line of work develops continual learning frameworks for VLAs which also  replay~\citep{stellarvla2026, ER-pretrainedVLAs, liu2026longlivedrobotscontinuallearning}. Across this line of work, forgetting is mitigated by storing real demonstrations, regularizing weights, or partitioning parameters. \methodabbrev is orthogonal: rather than maintaining a growing buffer of real trajectories, it leverages the world-action model itself to simulate past tasks, providing replay examples without retaining any real demonstrations.

\paragraph{Generative Replay} Unlike replay-based methods, generative replay synthesizes pseudo-samples for rehearsal~\citep{shin2017continuallearningdeepgenerative}. In robotics, CRIL~\citep{gao2021cril}, t-DGR~\citep{t-DGR}, and~\citep{pan2025continual} train generative models on past data to prevent forgetting. In contrast, \methodabbrev generates trajectories from current task data to facilitate continual learning. \methodabbrev shares this current-data insight but targets continual imitation learning. By leveraging a unified world-action model~\citep{cosmos-policy} for both action and future-frame prediction, \methodabbrev avoids model-based planning and eliminates the need for separate dynamics models.

\paragraph{World Models for robot control}
World models learn predictive representations of environment dynamics, originally proposed for sample-efficient reinforcement learning (RL)~\citep{ha2018worldmodels} and popularized by the Dreamer framework~\citep{hafner2020dreamcontrollearningbehaviors, dreamerv3}. While traditionally used for RL, the recent scaling of internet-level video data has enabled a new class of world-action models (WAMs) for imitation learning~\citep{cosmos-policy,lingbot-va2026,agibotworldcontributors2025agibotworldcolosseolargescale,ye2026dreamzero,zhu2025unifiedworldmodelscoupling,motubrainteam2026motubrainadvancedworldaction}. WAMs leverage unified architectures to jointly predict actions and future observations. While these models internalize physical causality, adapting such high-capacity architectures to non-stationary task streams remains an open continual learning challenge. 

% Preliminaries
%% World Action Models -- Just to introduce notation, no need to introduce tasks, just the pairs that tasks contain.

% Problem Formulation
%% Just our continual learning setting. We have a base policy trained on some tasks with some "true" data, want a policy trained on some new tasks WITHOUT using any of the "true" data. Emphasize with 1 sentence that existing replay based methods only use true replays, but its unrealistic and we dont want to.
%% This section will convey the "Base Stage" and "Continual Learning Stage" in the methods section v1

\vspace{-10pt}

\section{Preliminaries: World Action Models}
\label{sec:prelim}
\vspace{-0.1in}

We consider robotic policies learned through imitation learning from expert demonstrations. For each task $\mathcal{T}^k$, specified by a natural language instruction $\ell^k$, we assume access to a distribution of demonstrations $\mathcal{D}^k = \{(\ell^k,\tau_i^k)\}_{i=1}^{N_k}$, where each trajectory is defined as $\tau=\{(\mathbf{o}_t,\mathbf{a}_t)\}_{t=1}^{T}$ with observation $\mathbf{o}_t$ and action $\mathbf{a}_t$ at time step $t$. Each observation consists of multi-view RGB images $\mathbf{I}_t^1,\ldots,\mathbf{I}_t^n$ and the robot proprioceptive state $\mathbf{q}_t$.

WAMs are policies that jointly model future actions and future visual observations conditioned on language instructions and current observation. Unlike conventional policies that predict only actions, WAMs build upon generative video foundation models~\citep{nvidia2025cosmospredict2,wan2025wan, seedance2026seedance,zheng2026opensora20trainingcommerciallevel}, enabling both control and explicit prediction of future scene dynamics. Formally, at each time step $t$, a WAM parameterized by $\theta$ models the joint conditional distribution
\begin{equation}
    (\tilde{\mathbf{a}}_{t:t+H},\;
    \tilde{\mathbf{o}}_{t+H},\;
    \tilde{r}_t)
    \sim
    \pi_\theta
    \left(
    \cdot
    \mid
    \mathbf{o}_{t},
    \ell
    \right),
\end{equation}
where $\tilde{\mathbf{a}}_{t:t+H}$ denotes a predicted action chunk of horizon $H$, $\tilde{\mathbf{o}}_{t+H}$ is the predicted future observation, and $\tilde{r}_t \in [0,1]$ estimates task progress. Specifically, $\tilde{r}_t$ predicts the terminal reward $R(\mathbf{o}_T,\mathbf{a}_T)$ from the current state and serves as a dense proxy for proximity to task completion. The model parameters $\theta$ are optimized over $\mathcal{D}$ using a combination of behavioral cloning losses for actions, generative objectives for future observations, and regression losses for reward prediction.

% World Action Models (WAMs) are an imitation learning  for learning robotic policies from visuomotor demonstrations of specific tasks. Each demonstration consists of a task instruction expressed in natural language, denoted by $\ell$, and an action-observation trajectory, denoted by $\tau$:
% \begin{equation}
%     \tau = \{(\mathbf{o}_t, \mathbf{a}_t)\}_{t=1}^{T}
% \end{equation}
% Here, $\mathbf{o}_t$ is the robot's visual observation at time $t$, $\mathbf{a}_t$ is the action, and $T$ is the trajectory length. 
% % This trajectory is paired with the task instruction to construct the demonstration:
% % \begin{equation}
% %     \mathcal{D} = (\tau, l).
% % \end{equation}

% World Action Models, denoted by $\mathcal{\pi}_\theta$ and parameterized by $\theta$, learn to model the conditional distribution over future actions and future visual observations.
% \begin{equation}
%     (\tilde{\mathbf{a}}_{t:t+H}, \tilde{\mathbf{o}}_{t+H})
%     \sim
%     \pi_\theta(\mathbf{a}_{t:t+H}, \mathbf{o}_{t+H} \mid \mathbf{o}_{t}, \ell),
% \end{equation}
% where $H$ is the action chunking horizon, $\tilde{\mathbf{a}}_{t:t+H}$ denotes the generated action chunk, and $\tilde{\mathbf{o}}_{t+H}$ denotes the generated future visual observations. Tildes indicates actions and observations that are generated by the WAM rather than observed in real demonstrations. 

\vspace{-10pt}
\section{Method}
\label{sec:method}
\vspace{-0.1in}
In this section, we first present the continual learning problem formulation and then introduce our proposed framework for continual adaptation of WAMs, \textbf{\methodname} (\textbf{\methodabbrev}).

\subsection{Problem Formulation}
\label{sec:problem_formulation}

We consider continual adaptation of a pretrained WAM policy. Specifically, let $\pi_{\theta_0}$ denote a base policy trained on a set of previously learned tasks
$\mathcal{T}_{\mathrm{prev}}
=
\{\mathcal{T}_1,\mathcal{T}_2,\ldots,\mathcal{T}_M\}$.
Our goal is to adapt $\pi_{\theta_0}$ to a novel task $\mathcal{T}_k$, where $k>M$, while preserving performance on all tasks in $\mathcal{T}_{\mathrm{prev}}$. This results into an updated policy $\pi_{\theta_k}$.

For each previous task $\mathcal{T}_i \in \mathcal{T}_{\mathrm{prev}}$, we assume access only to the task-level language instruction $\ell_i$; no action-observation trajectories from previous tasks are retained. In contrast, for the current task $\mathcal{T}_k$, we assume access to both the task instruction $\ell_k$ and a distribution of expert demonstrations $\mathcal{D}_k
=
\{(\ell_k,\tau_k^n)\}_{n=1}^{N_k}$,
where $\tau_k^n$ denotes the $n$-th demonstration trajectory for task $\mathcal{T}_k$. Consequently, continual learning must be performed using demonstrations from the current task alone, while relying only on language instructions to preserve previously acquired behaviors.

A straightforward approach is to fine-tune the policy solely on $\mathcal{D}_k$. However, because this objective contains no explicit information about previous tasks, it can lead to catastrophic forgetting. Replay-based continual learning methods mitigate this issue by jointly training on stored demonstrations from prior tasks and current-task data. Although such approaches are infeasible in our setting due to the absence of previous trajectories, WAMs provide a unique advantage in this regime. Since WAMs jointly model future actions and future visual observations, they expose a generative interface capable of simulating prior task dynamics conditioned only on task instructions. We leverage this property to enable continual robot learning without storing or replaying ground-truth demonstrations from previous tasks.

% Our continual learning setting consists of two phases: \textit{(1) multi-task base stage} and \textit{(2) continual learning stage} following ~\citep{wan2024lotuscontinualimitationlearning,yu2026lifelongimitationlearningmultimodal}.

% \textbf{Base Stage.} In the base stage, policy is learned using a fixed set of $M$ tasks $\mathcal{T}_{\text{base}} = \{T_1, T_2, \ldots, T_M\}$ where each task consists of N demonstrations $\mathcal{D}_{\text{base}} = \{\mathcal{D}_1,
% \ldots, \mathcal{D}_M\}$. This stage equips the WAM with a broad manipulation prior over diverse skills, serving as the initialization for the subsequent continual learning stage.

% \textbf{Continual Learning Stage.} After Base stage, an agent is trained via imitation learning over a sequence of tasks
% presented one at a time. After base training, the agent encounters $K$ previously
% unseen tasks $\mathcal{T}_{\text{CL}} = \{T_M, T_{M+1}, \ldots, T_{M+K}\}$, with
% $\mathcal{T}_{\text{CL}} \cap \mathcal{T}_{\text{base}} = \emptyset$. At task $k$, the agent has
% access to a set of $N_k$ expert demonstrations for the current task:
%  $\mathcal{D}_k = \{\tau_{k,i}\}_{i=1}^{N_k}$, where each trajectory         
%  $\tau_{k,i} = \{(o_t, a_t)\}_{t=0}^{H_i}$,
% where crucially, demonstrations from previous tasks $\{D_1, \ldots, \mathcal{D}_{k-1}\}$ are \emph{no longer available}, reflecting a practical continual learning constraint where past data cannot be stored indefinitely.

\subsection{\methodname}
\label{sec:Regen}

\begin{figure}[h!]
    \centering
    \scalebox{0.9}{
    \includegraphics[width=\linewidth]{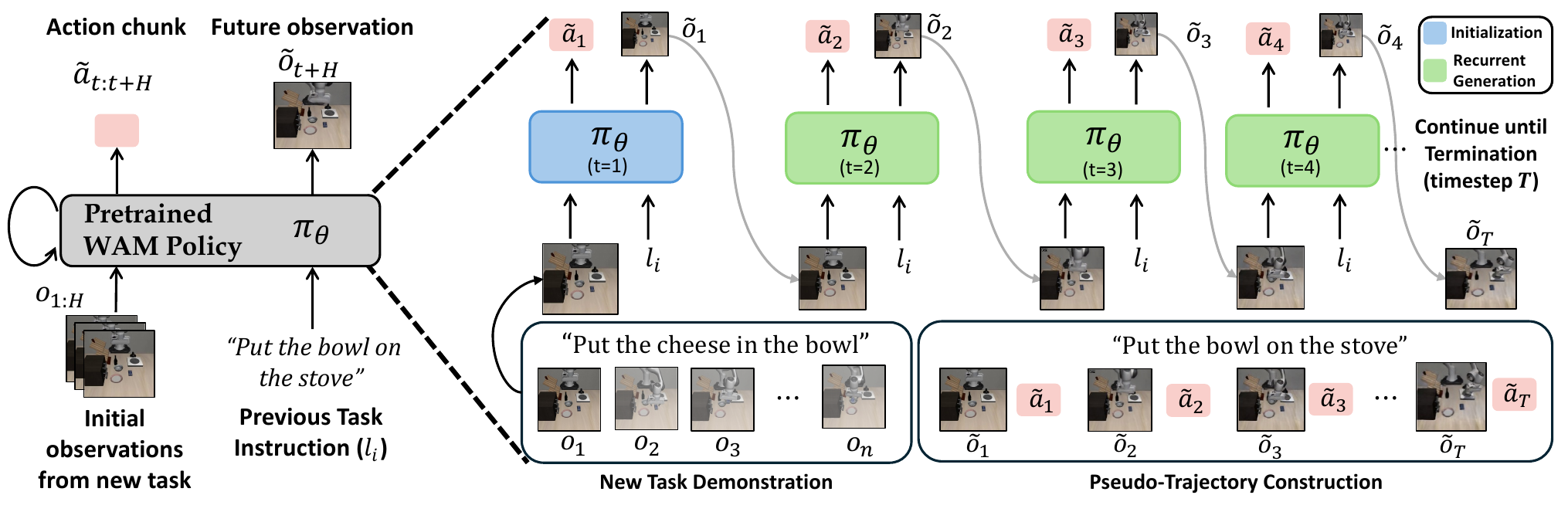}}
    \caption{\textbf{Overview of the pseudo-trajectory generation process.} \emph{Left:} unrolled view of the ~\methodabbrev. \emph{Right:} \methodabbrev rolls out $\pi_\theta$ recurrently to construct a pseudo-trajectory for a previous task. The policy is seeded with new-task observations and the previous task's instruction $\ell_i$ (initialization, blue), then each generated observation is fed back to produce the next $(\tilde{o}_t,\tilde{a}_t)$ pair (recurrent generation, green) until timestep $T$. For simplicity, we illustrate the process with $H=1$.} \vspace{-0.1in}
    \label{fig:Recurrent generation}
\end{figure}

Given the current task $\mathcal{T}_k$ with demonstrations $\mathcal{D}_k$ and a pretrained policy $\pi_{\theta}$, \methodabbrev{} generates pseudo-demonstrations for each previous task $\mathcal{T}_i$ ($i \leq M$) by conditioning $\pi_{\theta}$ on the corresponding task instruction $\ell_i$ and initializing the rollout from a real observation sampled from $\mathcal{T}_k$, as illustrated in Figure~\ref{fig:Recurrent generation}. At each rollout step, the model predicts an action chunk together with a future observation, which are recurrently fed back into the model to synthesize a pseudo-demonstration trajectory. Overall, pseudo-demonstration generation proceeds in three stages: initialization, recurrent generation, and termination, described below.

\textbf{Initialization Phase ($0 \leq t < H$).}
The rollout is initialized using one full action chunk from the current-task demonstrations, since at least one real observation context is required before recursive generation can begin. During this phase, the model conditions on real observations sampled from $\mathcal{D}_k$, i.e., $\mathbf{o}^{\mathrm{in}}_t = \mathbf{o}_t$ for $0 \leq t < H$.

\textbf{Recurrent Generation Phase ($H \leq t \leq T_{\max}$).}
After initialization, real observations are no longer used. Instead, the model recursively conditions on its own previously generated observations to produce a fully generative rollout:
\begin{equation} \mathbf{o}^{\text{in}}_t = \begin{cases} \mathbf{o}_t, & 0 \leq t < H, \\[6pt] \tilde{\mathbf{o}}_{t}\;\;\text{where}\;\; (\tilde{\mathbf{a}}_{t-H:t},\, \tilde{\mathbf{o}}_{t}) \sim \pi_{\theta}\!\left(\cdot \mid \mathbf{o}^{\text{in}}_{t-H},\, \ell_i\right), & H \leq t \leq T_{\max}. \end{cases} \label{eq:recurrence} \end{equation}
This recursive feedback of generated observations induces a recurrent rollout process, enabling reconstruction of prior-task trajectories without storing any previous task demonstrations.

\textbf{Termination.}
Trajectory generation terminates either at a maximum horizon $T_{\max}$ or earlier when the goal-reward head consistently predicts task completion. Specifically, generation stops when the predicted goal reward $\tilde{r}_t$ exceeds $0.99$ for three consecutive rollout steps and attains $1.0$ at least once within that interval.

\textbf{Pseudo-Trajectory Construction.}
At each rollout step, the input observation is paired with the first action of the predicted action chunk to construct a pseudo-trajectory:
\begin{equation}
    \tilde{\tau}^{\,i}
    =
    \left\{
    \left(
    \mathbf{o}^{\mathrm{in}}_t,
    \tilde{\mathbf{a}}_t
    \right)
    \right\}_{t=0}^{T_i},
\end{equation}
where $\tilde{\mathbf{a}}_t$ denotes the first action of the chunk predicted at time step $t$, and $T_i \leq T_{\max}$ is the generated trajectory length for task $\mathcal{T}_i$. The pseudo-demonstration set is formed by aggregating generated rollouts across all previous tasks, i.e., $\mathcal{R}_k = \bigcup_{i=1}^{M}\tilde{\tau}^{\,i}$. The policy is then updated via behavioral cloning using the combined training set $\mathcal{D}^{+}_k = \mathcal{D}_k \cup \mathcal{R}_k$.

\textbf{Training Objective.}
The WAM policy is trained on the union of current-task demonstrations and replayed pseudo-trajectories, $\mathcal{D}_k^{+}$, in order to acquire the new task while retaining performance on previous tasks:
\begin{equation}
    \min_{\theta}
    \;
    \mathbb{E}_{(\mathbf{o}_t,\mathbf{a}_t,\ell)\sim\mathcal{D}_k^{+}}
    \left[
    \mathcal{L}_{\mathrm{BC}}
    \left(
    \pi_{\theta}(\mathbf{o}_t,\ell),
    \mathbf{a}_t
    \right)
    \right],
    \label{eq:objective}
\end{equation}
where $\mathcal{L}_{\mathrm{BC}}$ denotes the behavioral cloning loss. The task instruction $\ell$ corresponds to $\ell_k$ for samples from the current-task dataset $\mathcal{D}_k$, and to $\ell_i$ for pseudo-trajectories generated for previous task $\mathcal{T}_i$. In this way, \methodabbrev{} mitigates catastrophic forgetting by approximating the trajectory distributions of previous tasks without storing ground-truth demonstrations.

% \Statex \textbf{Phase 2: Continual Policy Update}
% \vspace{-1pt}

% \For{each training iteration}
%     \State Sample current batch $(o_{\leq t}, a_t) \sim \mathcal{D}_k$
%     \State Sample replay batch $(\hat{o}_{\leq t}, \hat{a}_t) \sim \mathcal{B}$
%     \State $\mathcal{L}_{\text{\methodabbrev}} \leftarrow
%            \mathcal{L}_{\text{BC}}(\W_{\theta_k}, \mathcal{D}_k) +
%            \lambda\; \mathcal{L}_{\text{BC}}(\W_{\theta_k}, \mathcal{B})$
%     \State Update $\W_{\theta_k}$ and $\Phi_\theta$ via
%            $\nabla_\theta \mathcal{L}_{\text{\methodabbrev}}$
% \EndFor

% \State \Return $\W_{\theta_k}$, $\Phi_\theta$, $\mathcal{B}$

%===============================================================================
\vspace{-10pt}
\section{Experimental Results}
\label{sec:exp}
%We design our experiments to answer the following questions: (1) Does \methodname mitigate catastrophic forgetting in the continual learning setting? (2) How does the quality of generated samples affect continual learning performance? (3) Does the pseudo-trajectories generated by \methodabbrev remain consistent with the actions?
We conduct experiments in both simulation and real-world environments to evaluate the effectiveness of the pseudo-trajectories generated by \methodabbrev. We further perform representation analyses to characterize the quality of the generated trajectories and assess their deviation from perfect demonstrations.
\vspace{-5pt}
\subsection{Implementation Details \& Evaluation Metrics}
\label{sec:implementation}

We instantiate Cosmos-Policy~\citep{cosmos-policy} as the underlying WAM, initialized from Cosmos-Predict2-2B~\citep{nvidia2025cosmospredict2}. The model conditions on third-person RGB and wrist-camera observations, the current proprioceptive state, and a language instruction, and jointly predicts an action chunk of horizon $H=16$ at each timestep. Unless otherwise specified, we adopt the training hyperparameters of Cosmos-Policy. The base policy is trained for $10$K iterations, and each continual learning stage fine-tunes the policy for $2$K iterations from the checkpoint of the previous stage. For replay generation, we synthesize $10$ pseudo-trajectories per previous task in $\mathcal{T}_{\mathrm{prev}}$. Additional architectural details, hyperparameters and ablations are provided in the Appendix.

For evaluation, we report three standard continual learning metrics: Forward Transfer (FWT)~\citep{liu2023libero,wan2024lotuscontinualimitationlearning}, Negative Backward Transfer (NBT)~\citep{ER-pretrainedVLAs}, and Area Under the Curve (AUC)~\citep{liu2023libero,wan2024lotuscontinualimitationlearning}. All metrics are computed using task success rates. Let $N$ denote the total number of tasks in the continual learning sequence, and let $r_{i,j}$ denote the success rate on task $j$ after training up to task $i$.
Then, FWT measures the ability to acquire new tasks, while NBT quantifies forgetting on previously learned tasks:

\[
\textbf{FWT}
=
\frac{1}{N}
\sum_{n=1}^{N}
r_{n,n},
\qquad
\textbf{NBT}
=
\frac{1}{N-1}
\sum_{n=1}^{N-1}
\text{NBT}_n,
\qquad
\text{NBT}_n
=
\frac{1}{N-n}
\sum_{p=n+1}^{N}
\left(
\frac{r_{n,n}-r_{p,n}}{r_{n,n}}
\right)
\]
Higher FWT indicates stronger forward transfer, whereas lower NBT corresponds to reduced forgetting.
Finally, AUC measures overall performance across both current and previously learned tasks:
\[
\textbf{AUC}
=
\frac{1}{N}
\sum_{n=1}^{N}
\frac{1}{N-n+1}
\left(
r_{n,n}
+
\sum_{p=n+1}^{N}
r_{p,n}
\right).
\]
Higher AUC indicates better overall continual learning performance.
\subsection{LIBERO Simulated Environment}

\textbf{Setting.}
We conduct our simulation experiments on LIBERO benchmark~\citep{liu2023libero}. We use three different task suites that capture distinct forms of distribution shift:(LIBERO-Spatial, LIBERO-Object, LIBERO-Goal). Each suite contains $10$ tasks, with $50$ high-quality human-teleoperated demonstrations per task. Our continual learning setup involves two stages, similar to~\citep{wan2024lotuscontinualimitationlearning,yu2026lifelongimitationlearningmultimodal} (1) a base stage and (2) a continual learning stage. In the base stage, six tasks are used for pretraining. During continual learning, the remaining four tasks are introduced sequentially, with one new task added at each continual learning stage. We follow the task ordering defined for each benchmark suite in~\citep{ER-pretrainedVLAs}.

\textbf{Baseline Methods.}
We compare \methodabbrev against the following continual learning approaches:
\begin{enumerate}
    \item \textbf{Sequential Fine-Tuning (Seq-FT)}~\citep{liu2023libero}: fine-tunes the base policy on each new task without any forgetting mitigation.
    \item \textbf{Sequential LoRA (Seq-LoRA)}: a parameter-efficient variant of Seq-FT using LoRA~\citep{hu2022lora} fine-tuning instead of full-model updates.
    \item \textbf{Elastic Weight Consolidation (EWC)}~\citep{kirkpatrick2017overcoming}: a regularization based approach which penalizes changes to parameters important for prior tasks.
    \item \textbf{PackNet}~\citep{mallya2018packnetaddingmultipletasks}: an iterative pruning approach that frees up redundant parameters after training each task and uses them to learn new tasks, keeping previously allocated parameters frozen.
    \item \textbf{Experience Replay (ER)}~\citep{liu2023libero}: a replay-based method that stores real demonstrations of  prior tasks and mixes them with current-task data during training. We include ER only as an upper-bound reference, since it violates our no-real-data assumption.
    \item \textbf{Rollouts-as-Replay (RAR)}: like \methodabbrev, stores no real trajectories; at each stage we roll out previous-task policies in the simulator and use the rollouts as replay data, isolating \emph{simulator-rendered} from \emph{model-generated} replay.
\end{enumerate}

\begin{table*}[h]\vspace{-0.1in}
\centering
\caption{Comparison of \methodabbrev with traditional continual learning methods on LIBERO benchmarks.} \vspace{-0.1in}
\label{tab:libero_results}

\begin{minipage}{0.55\linewidth}
\centering
\resizebox{\linewidth}{!}{
\begin{tabular}{@{}l ccc ccc@{}}
\toprule
& \multicolumn{3}{c}{LIBERO-Object} & \multicolumn{3}{c}{LIBERO-Goal} \\
\cmidrule(lr){2-4} \cmidrule(lr){5-7}
Method & FWT $\uparrow$ & NBT $\downarrow$ & AUC $\uparrow$
& FWT $\uparrow$ & NBT $\downarrow$ & AUC $\uparrow$ \\
\midrule
Seq-FT~\citep{liu2023libero} & 92.7 & 82.6 & 24.9 & 90.6 & 100 & 10.3 \\
Seq-LoRA~\citep{lee2024incremental} & 93.5 & 99.9 & 11.1 & 83  & 99.6  &  9.3    \\
EWC~\citep{edwards2019imitatinglatentpoliciesobservation} & 94.9 & 87.5 & 25.1 & 83.6 & 99.4  & 9.9  \\
PackNet~\citep{mallya2018packnetaddingmultipletasks} & 95.5 & 99.7 & 11.5 & 92 & 100 & 10.5\\
\rowcolor{lightgray}
\midrule
ER~\citep{ER} & 95.7 & 4.8 & 93.4 & \textbf{94.0} & \textbf{7.2} & \textbf{92.4} \\
RAR & \textbf{96.9} & \textbf{3.0} & \textbf{95.2} & 92.8 & 15.4 & 82.6 \\
\textbf{\methodabbrev (ours)} & 95.3 & 26.1 & 65.5 & 90.6 & 44.9 & 40.8 \\
\bottomrule
\end{tabular}
}
\end{minipage}
\hspace{0.015\linewidth}
\begin{minipage}{0.375\linewidth}
\centering
\resizebox{\linewidth}{!}{
\begin{tabular}{@{}l ccc@{}}
\toprule
& \multicolumn{3}{c}{LIBERO-Spatial} \\
\cmidrule(lr){2-4}
Method & FWT $\uparrow$ & NBT $\downarrow$ & AUC $\uparrow$ \\
\midrule
Seq-FT~\citep{liu2023libero} & \textbf{87.4} & 99.8 & 10.8 \\
Seq-LoRA~\citep{lee2024incremental} & 80 & 99.6 & 8.9 \\
EWC~\citep{edwards2019imitatinglatentpoliciesobservation} & 86.4 & 99.9 & 10.3 \\
PackNet~\citep{mallya2018packnetaddingmultipletasks} & 87.2 & 100 & 10.4  \\
\rowcolor{lightgray}
\midrule
ER~\citep{ER} & 86.4 & \textbf{-0.28} & \textbf{87.8} \\
RAR & 87 & -0.02 & 85.8 \\ 

\textbf{\methodabbrev$^{\dagger}$ (ours)} & 87.2 & 17.6 & 76.9 \\
\bottomrule
\end{tabular}
}
\end{minipage}
\end{table*}
   
\textbf{Results.} Table~\ref{tab:libero_results} reports continual learning performance on the three LIBERO benchmark suites. Firstly, Sequential finetuning (Seq-FT) of WAMs achieves strong forward transfer but suffers from near-complete forgetting of previously learned tasks, consistent with prior observations in behavior cloning approaches~\citep{liu2023libero,wan2024lotuscontinualimitationlearning,clare}. While Seq-FT fully adapts to new tasks but forgets prior ones, the non-replay baselines (Seq-LoRA, EWC, and PackNet) retain partial knowledge of previous tasks but cannot perform them reliably. Next, ER yields the strongest overall retention and highest AUC, confirming that replaying real trajectories remains the most effective strategy for mitigating forgetting. However, we treat ER as a reference rather than a directly comparable baseline, since it assumes access to stored demonstrations from previous tasks, which violates the constraints of our setting.
RAR mitigates forgetting by replaying simulator-collected rollouts, achieving performance close to ER with minimal forgetting and without storing real replay buffers. While effective in simulation, such approaches are impractical in real-world robotics, where replay generation requires re-deploying the robot and re-interacting with the environment to collect trajectories.

In contrast, \methodabbrev~substantially reduces forgetting using only pseudo-trajectories generated by the WAM itself, without storing any real demonstrations. Compared to sequential fine-tuning, replaying pseudo-trajectories preserves forward transfer while significantly improving retention, reducing NBT by more than $50\%$ across multiple suites. These results demonstrate that the generative capabilities of WAMs can serve as an effective form of implicit memory for continual robot learning.
For LIBERO-Spatial, we introduce a variant denoted as \textbf{\methodabbrev$^{\dagger}$}, where replay generation is initialized using object configurations sampled from previous tasks. This modification is necessary because the benchmark evaluates spatial generalization across object arrangements, and replay generation requires the corresponding objects to be present in the scene. For example, trajectories involving an object absent from the current environment cannot be reliably synthesized.

\textbf{VLA vs.\ WAM Continual Learning.}
Compared to pretrained VLAs such as $\pi_{0.5}$~\citep{pi05}, forgetting in WAMs is initially more severe; however, Table~\ref{tab:libero_goal} shows that large-scale pretraining alone does not eliminate catastrophic forgetting. Despite being trained on substantially larger robotic datasets~\citep{open_x_embodiment_rt_x_2023}, $\pi_{0.5}$ still exhibits significant degradation on previously learned tasks during continual adaptation. In contrast, \methodabbrev{} achieves stronger retention using only generated pseudo-replays, by leveraging the joint action-observation generative modeling capability unique to WAMs. Unlike standard VLAs~\citep{kim_openvla_2024,pi05,bu_univla_2025}, WAMs can explicitly generate future observations, enabling replay without additional data collection or storage. These results suggest that WAMs provide a more suitable foundation for lifelong robot learning.

%The gap between RAR and \methodabbrev quantifies the cost incurred by replaying \emph{generated} rather than \emph{simulator-rendered} rollouts, and points to the visual fidelity of generated trajectories (Fig.~\ref{fig:quantifying_plot}, left) as the primary challenge for further closing this gap.

\begin{table*}[t!]
\centering
\begin{minipage}{0.45\linewidth}
\centering
\captionof{table}{VLA vs WAM: continual learning performance on LIBERO-Goal.}
\label{tab:libero_goal}
\scalebox{0.75}{
\begin{tabular}{@{}ll ccc@{}}
\toprule
\cmidrule(lr){3-5}
Policy & Method & FWT $\uparrow$ & NBT $\downarrow$ & AUC $\uparrow$ \\
\midrule
\multirow{1}{*}{$\pi_{0.5}$ ~\citep{pi05}}
& Seq-FT & \textbf{96.8} & 88 & 35.5 \\
\midrule
\multirow{3}{*}{Cosmos-Policy~\citep{cosmos-policy}}
& Seq-FT  & 90.6 & 100  & 10.3   \\
& \textbf{\methodabbrev (ours)} & 90.6 & \textbf{38.7} & \textbf{40.8} \\
\bottomrule
\end{tabular}
}
\end{minipage}
\hfill
\begin{minipage}{0.45\linewidth}
\centering
\captionof{table}{Results of \methodabbrev~in real-world manipulation environment.}
\label{tab:realworld}
\begin{tabular}{@{}l ccc@{}}
\toprule
Method & FWT $\uparrow$ & NBT $\downarrow$ & AUC $\uparrow$ \\
\midrule
Seq-FT   & 50   & 96.3    & 13.8  \\
\textbf{\methodabbrev (ours)}  & \textbf{80}  & \textbf{60.5} & \textbf{53.8} \\
\bottomrule
\end{tabular}
\end{minipage} \vspace{-0.2in}
\end{table*}

\subsection{Real-world Single-arm Manipulation}

\begin{wrapfigure}{h}{0.5\linewidth}

    \centering
    \scalebox{0.8}{
    \includegraphics[width=0.8\linewidth]{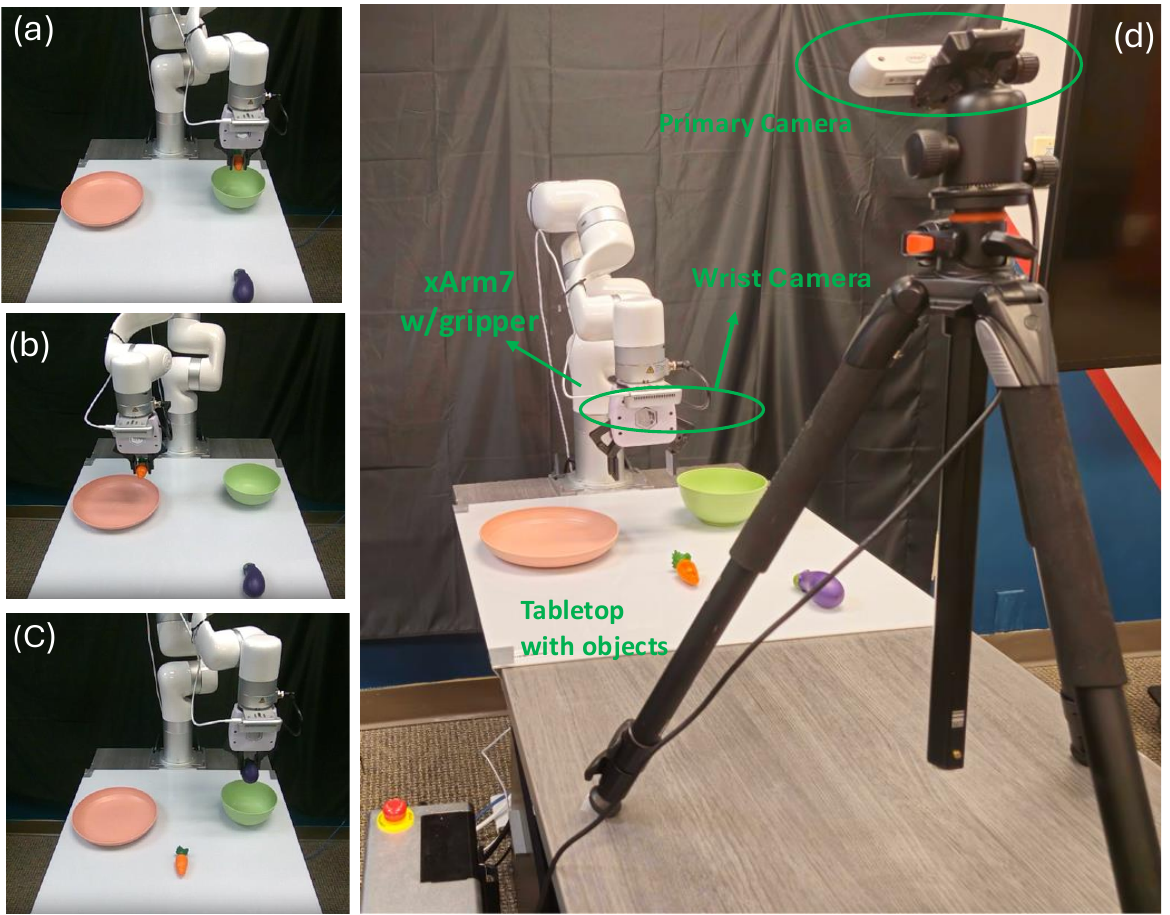}}
    \caption{\textbf{Real-world setting.} a), b), c) illustrate tasks T1, T2, and T3, respectively. d) Our real-robot setup consists of an xArm7 manipulator, a wrist-mounted gripper camera, and a third-person RGB-D camera.} 
    \label{fig:real_robot_setup}
\end{wrapfigure}

\textbf{Setting.} We conduct all real-world experiments on an xArm7 robotic manipulator. Our setup is shown in Figure~\ref{fig:real_robot_setup}.
We evaluate \methodabbrev on three real-world manipulation tasks with a shared pick-and-place structure but different object-goal combinations:
\textit{(T1) Put carrot in bowl:} place the carrot inside the bowl;
\textit{(T2) Put carrot on plate:} place the carrot on the plate; and
\textit{(T3) Put eggplant in bowl:} place the eggplant inside the bowl.
Tasks are introduced sequentially in the order $\mathrm{T1} \rightarrow \mathrm{T2} \rightarrow \mathrm{T3}$, with continual adaptation performed between stages.
We collect $50$ teleoperated demonstrations per task at a control frequency of $15$ Hz. Evaluation metrics are averaged across the two continual learning stages, and each policy is evaluated over $10$ randomized trials with varying object placements and initial gripper configurations.
\textbf{Results.}
Table~\ref{tab:realworld} reports the real-world continual learning results. \methodabbrev substantially outperforms sequential fine-tuning, reducing NBT from $96.3$ to $60.5$ (approximately $40\%$ less forgetting) while improving FWT from $50$ to $80$. We attribute the improved forward transfer to the regularizing effect of replayed pseudo-trajectories, particularly in the low-data regime where the base policy is initialized from a single task and therefore exhibits a limited prior over manipulation behaviors. These results demonstrate that \methodabbrev extends effectively beyond simulation and provides a practical approach for continual learning on real robotic systems.

\subsection{\methodabbrev Analyses}
We further analyze how \methodabbrev preserves the policy's internal representations and behaviors throughout continual learning and study two key design choices in pseudo-trajectory generation.

\begin{figure}[h!]
    \centering
    \scalebox{1}{
    \begin{subfigure}[t]{0.45\textwidth}
        \centering
        \includegraphics[width=\textwidth]{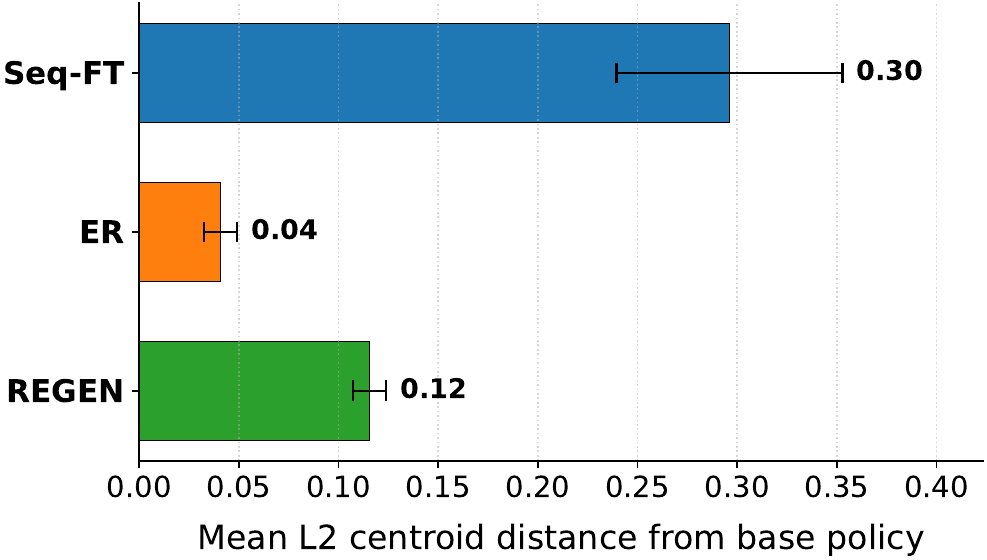}
        \caption{}
        \label{fig:rep_drift}
    \end{subfigure}
    \hfill
    \begin{subfigure}[t]{0.5\textwidth}
        \centering
        \includegraphics[width=\textwidth]{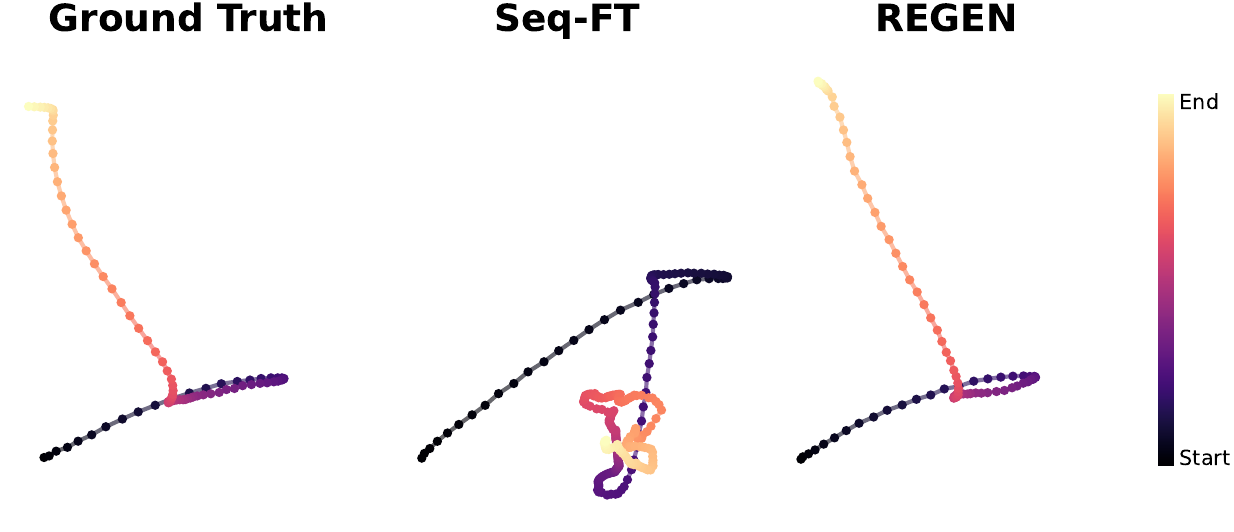}
        \caption{}
        \label{fig:trajectory_plot}
    \end{subfigure}}
    \vspace{-0.2cm}
    \caption{\textbf{(a)} Action representation drift from the base policy after the first continual learning stage between Seq-FT, ER, and \methodabbrev \textbf{(b)} XY-projection of trajectories predicted by Seq-FT and \methodabbrev on a previously seen task, compared with the ground-truth demonstration.}
    \label{fig:analysis_plot}
\end{figure}

\textbf{Action representation drift.}
Figure~\ref{fig:rep_drift} measures the drift in action representations after the first continual learning stage. Across six base-stage tasks, we compute the centroid of the action latent representations under both the base policy and the continually adapted policy, and report the mean $\ell_2$ distance between them. Seq-FT exhibits substantial representation drift (up to $0.3$), consistent with its severe catastrophic forgetting. In contrast, both ER ($ 0.04$) and \methodabbrev ($0.12$) maintain significantly lower drift despite \methodabbrev relying solely on generated pseudo-trajectories. These results suggest that \methodabbrev effectively preserves the action representations learned by the base policy.

\textbf{Visualization of predicted actions.}
In Fig.~\ref{fig:trajectory_plot}, we additionally compare action trajectories generated by Seq-FT and \methodabbrev against ground-truth demonstrations on a previously learned evaluation task by projecting trajectories onto the XY plane. Seq-FT produces trajectories that deviate substantially from the ground-truth behavior, exhibiting erratic motion patterns indicative of catastrophic forgetting. In contrast, trajectories generated by \methodabbrev closely match the ground-truth trajectory in both shape and temporal progression, indicating that the policy retains the underlying structure of previously acquired manipulation skills.
\vspace{-0.5em}
\paragraph{Number of replays.} We study the effect of the number of pseudo-replays from ~\methodabbrev on continual learning performance. On LIBERO-Object (Tab.~\ref{tab:num_replays}), we find that $10$ replay examples per task achieve lower NBT than $5$ replays while maintaining comparable FWT and AUC, indicating that additional replay diversity improves backward transfer.
\vspace{-0.5em}
\paragraph{Termination criterion.} We compare two stopping criteria for pseudo-trajectory generation: a fixed-horizon rule that always rolls out to a maximum of $H$ steps, and a goal-reward-based rule that terminates when a function defined on the WAM's predicted value indicates task success. Table~\ref{tab:termination} reports the PSNR values for trajectories generated under both rules on LIBERO-Goal. The goal-reward-based rule yields higher-fidelity pseudo-trajectories (PSNR $20.3$), since early termination avoids extending the rollout into low-quality frames generated through recursive prediction. We therefore adopt the goal-reward criterion in \methodabbrev. Specifically, the reward function operates over a sliding window of 3 consecutive value predictions and signals task success when at least one prediction reaches $1.0$ and the remaining exceed $0.99$.

\begin{table}[h!]
\centering
\begin{minipage}{0.45\linewidth}
    \centering
    \captionof{table}{Effect of the number of pseudo-replays on LIBERO-Object.}
    \label{tab:num_replays}
    \resizebox{\linewidth}{!}{
    \begin{tabular}{@{}l ccc@{}}
    \toprule
    \# No. of Replays & FWT $\uparrow$ & NBT $\downarrow$ & AUC $\uparrow$ \\
    \midrule
    5  & 96.5 & 33 & 62.5 \\
    10 & 96.1 & 31 & 62.3 \\
    \bottomrule
    \end{tabular}
    }
\end{minipage}
\hfill
\begin{minipage}{0.4\linewidth}
    \centering
    \captionof{table}{Effect of the termination criterion in ~\methodabbrev.}
    \label{tab:termination}
    \resizebox{\linewidth}{!}{
    \begin{tabular}{@{}lccc@{}}
    \toprule
    Termination criterion & PSNR ($\uparrow$) \\
    \midrule
    Fixed Horizon with H=200 & 18.4  \\
    Fixed Horizon with H=150 & 19.5  \\
    Goal-reward function  & 20.3  \\
    \bottomrule
    \end{tabular}
    }
\end{minipage}
\end{table}

\begin{figure}[h!] \vspace{-0.1in}
    \centering
    \scalebox{0.9}{
    \includegraphics[width=1\linewidth]{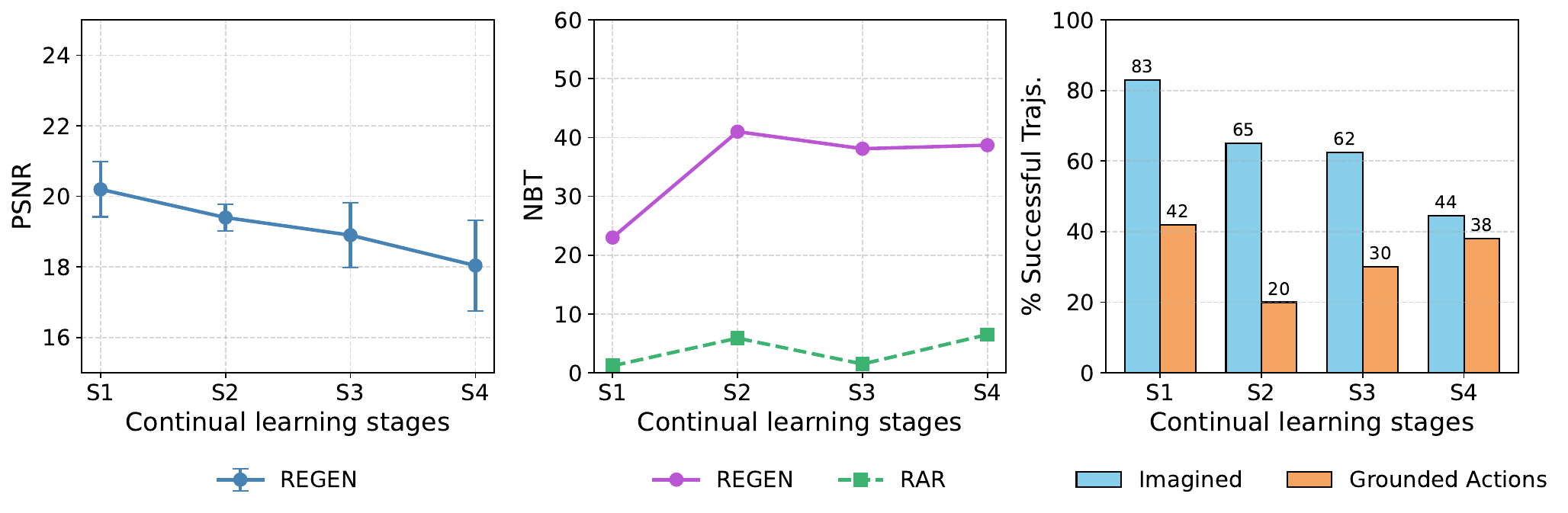}} \vspace{-0.1in}
    \caption{\textbf{(Left)} PSNR (with std) of generated trajectories from \methodabbrev across CL stages on LIBERO-Goal. \textbf{(Middle)} NBT comparison between \methodabbrev and RAR. \textbf{(Right)} successful imagined trajectories vs.\ successful action grounded trajectories on LIBERO-Goal.}\vspace{-0.4cm}
\label{fig:quantifying_plot}
\end{figure} 

\vspace{-0.1in}
\section{Limitations of \methodabbrev \& Future Direction}\vspace{-0.1in}
\label{challenges}

Although \methodabbrev substantially mitigates catastrophic forgetting without access to previous-task data, a performance gap remains relative to the privileged ER baseline. We identify the primary bottleneck as the limited generative fidelity of current WAMs.

\textbf{Visual fidelity of generated observations.}
The first limitation concerns the quality of pseudo-trajectories generated across successive continual learning stages. Figure~\ref{fig:quantifying_plot} (Left) shows a monotonic decline in the PSNR of synthesized trajectories as continual adaptation progresses, with qualitative examples of this degradation across stages shown in Appendix~\ref{app:obs_quality}. We attribute this degradation to two compounding factors: (i) visual artifacts and blurriness in generated observations that are recursively reused as conditioning inputs for future prediction, and (ii) accumulation of errors as the model is repeatedly updated over continual learning stages. Importantly, the reduction in PSNR correlates with increased NBT, suggesting that lower-fidelity pseudo-trajectories provide progressively weaker supervisory signals for retaining previously learned behaviors.
This observation is further supported by the comparison with RAR. Replacing simulator-generated replay trajectories with WAM-generated pseudo-trajectories leads to a substantial drop in continual learning performance (Fig.~\ref{fig:quantifying_plot}, middle), indicating that generative fidelity is a key limiting factor for replay quality.  

\textbf{Mismatch between predicted states and actions.}
A second limitation arises from inconsistencies between generated future observations and the corresponding predicted actions. Figure~\ref{fig:quantifying_plot} (Right) compares the \emph{imagined} success rate, evaluated from generated observations, with the \emph{grounded} success rate obtained by executing the predicted actions in the simulator. At Stage~1, the model achieves an imagined success rate of $83\%$ but only a grounded success rate of $42\%$. This discrepancy indicates that the WAM frequently predicts visually plausible successful outcomes while generating actions that are physically insufficient to accomplish the task (See Appendix~\ref{app:action_obs_mismatch}). We hypothesize that this decoupling stems from degradation in generated observations, which weakens the alignment between visual predictions and action dynamics. Consequently, improving the generative consistency and visual fidelity of WAMs remains a future direction for narrowing the performance gap to experience replay.

%==========================================================
\section{Conclusion}

\label{sec:conclusion}
We presented Recurrent Generative Replay (\methodabbrev), the first continual learning framework to leverage the generative capabilities of WAMs as a native replay mechanism. By synthesizing pseudo-demonstrations conditioned on prior task instructions, current visual observations, and its own generated observations, \methodabbrev mitigates catastrophic forgetting without requiring access to stored demonstrations from previous tasks. Experiments in both simulation and real-world manipulation demonstrate its effectiveness for continual robot learning. Finally, our analysis identifies long-horizon degradation in generated observations as the primary bottleneck, highlighting an important direction for future advances in WAMs and generative replay.
%Across simulated and real-world manipulation settings, \methodabbrev reduces forgetting by up to 50\% relative to naive sequential fine-tuning, approaching the performance of methods with privileged access to real replay data. %Our analysis further identifies the key bottlenecks limiting \methodabbrev: long-horizon visual degradation and action-observation inconsistency in WAM-generated rollouts. As WAMs continue to mature, we expect these limitations to diminish, and with them, the remaining gap between generative and real experience replay.

%\section{Limitations}

%Our method is limited to same environment. Novel scene generation : If the  previous task is in a completely different environment from current Task

%===============================================================================

\clearpage
% The acknowledgments are automatically included only in the final and preprint versions of the paper.
\acknowledgments{This work was supported in part by the National Science Foundation (IIS-2245652) and the University of North Carolina at Charlotte. Computational resources were provided by the NSF National AI Research Resource Pilot (NAIRR240338) and NCShare.}

%===============================================================================

% no \bibliographystyle is required, since the corl style is automatically used.
\bibliography{main,unilact}  % .bib

\appendix

\section*{Appendix}

% The appendix provides additional details supporting the main paper:
% \begin{itemize}
%     \setlength\itemsep{2pt}
%     \item \textbf{Section~\ref{app:implementation}} : Implementation details, including model architecture, and pseudo-code for \methodabbrev.
%     \item \textbf{Section~\ref{app:training}} : Training details, hyperparameters and evaluation protocol.
%     \item \textbf{Section~\ref{app:additional_baselines}} : Additional continual learning baselines.
%     \item \textbf{Section~\ref{app:qualitative-examples}} : Examples of Psuedo-trajectories generated in \methodabbrev and Qualitative results of \methodname.
% \end{itemize}

\section{Implementation Details}
\label{app:implementation}
We provide the implementation details of the base WAM and the \methodabbrev algorithm.
\subsection{Base WAM implementation}
In \methodabbrev, we use Cosmos-Policy~\citep{cosmos-policy} as our WAM, initialized from Cosmos-Predict2-2B weights~\citep{nvidia2025cosmospredict2}. Cosmos-Policy is built on a latent video diffusion model that, conditioned on the current observation (primary RGB image, wrist RGB image, and robot proprioceptive state) and a natural-language task instruction, jointly predicts an action chunk, future observations, and a reward value. Visual observations are encoded with the Wan2.1 spatiotemporal VAE tokenizer~\citep{wan2025wan}, and language instructions are encoded with a pretrained T5-XXL encoder~\citep{raffel2023exploringlimitstransferlearning}. All actions and proprioceptive states are normalized to $[-1, +1]$ before being converted into latent frames. The model is trained to jointly denoise the action chunk, future observation, and reward value latents under a flow-matching diffusion objective~\citep{karras2022elucidatingdesignspacediffusionbased}. We follow the policy model training strategy of Cosmos-Policy.

\subsection{Pseudo code of \methodabbrev }

Algorithm~\ref{alg:rgr} summarizes \methodabbrev: In each continual learning stage, we generate pseudo-trajectories for every previous task by recurrently rolling out the current WAM policy, then update the policy on a mixture of pseudo-trajectories and new-task demonstrations.

\begin{algorithm}[h!]
\caption{\methodname{} (\methodabbrev): Pseudo-Trajectory Generation and policy update during continual learning}
\label{alg:rgr}
\scalebox{0.8}{%
\begin{minipage}{\linewidth}
\begin{algorithmic}[1]
\Require Current task $\mathcal{T}_k$, current-task demonstrations $\mathcal{D}_k$, pretrained policy $\pi_{\theta}$, previous-task instructions $\{\ell_i\}_{i=1}^{M}$, chunk length $H$, max horizon $T_{\max}$, goal-reward threshold $\delta$, replays per task $N$, number of training iterations $I$, Trajectory Termination function \textsc{Terminate}
\State $\mathcal{R}_k \leftarrow \emptyset$ \Comment{Initialize pseudo-trajectories set }
\For{each previous task $\mathcal{T}_i$, $i = 1, \ldots, M$}
    \For{$n = 1$ to $N$}
        \State $\tilde{\tau} \leftarrow \emptyset$, \quad $t \leftarrow 0$
        \State $V_{\text{win}} \leftarrow$ empty deque of size $3$
        \While{$t < T_{\max}$ \textbf{ and not} \textsc{Terminate} ($V_{\text{win}}$, $\delta$)}
            \If{$t < H$}
                \State $\mathbf{o}^{\text{in}} \leftarrow \mathbf{o}_t$
                       \Comment{Real observation from $\mathcal{D}_k$}
            \Else
                \State $\mathbf{o}^{\text{in}} \leftarrow \tilde{\mathbf{o}}_{t}$
                       \Comment{Predicted observation from WAM}
            \EndIf
            \State $(\tilde{\mathbf{a}}_{t:t+H},\; \tilde{\mathbf{o}}_{t+H},\; \tilde{v}_t) \sim \pi_{\theta}\!\left(\cdot \mid \mathbf{o}^{\text{in}},\, \ell_i\right)$
            \State $\tilde{\tau} \leftarrow \tilde{\tau} \cup \{(\mathbf{o}^{\text{in}},\, \tilde{\mathbf{a}}_t)\}$
            \State append $\tilde{v}_t$ to $V_{\text{win}}$
            \State $t \leftarrow t + 1$
        \EndWhile
        \State $\mathcal{R}_k \leftarrow \mathcal{R}_k \cup \{\tilde{\tau}\}$
    \EndFor
\EndFor
\State \textbf{// Policy update with mixed data ($\mathcal{D}^{+}_k = \mathcal{D}_k \cup \mathcal{R}_k$})
\For{iteration $= 1$ to $I$}
    \State Sample mini-batch from $\mathcal{D}^{+}_k$
    \State $\theta \leftarrow \theta - \eta \, \nabla_{\theta} \mathcal{L}(\theta)$ 
\EndFor
\State \Return updated policy $\pi_{\theta}$
\end{algorithmic}
\end{minipage}%
}
\end{algorithm}
\section{Training Details}
\label{app:training}

In this section, we detail the training settings for  continual learning in both simulation (i.e. LIBERO [38]) and real-world environments.
\subsection{Our Continual Learning Setting}
\textbf{Base stage.}  In LIBERO, the base stage consists of the six tasks of each suite where as in real-world setting, the base stage includes only one task.

\textbf{Continual learning stage.} In each subsequent stage, the policy is  fine-tuned from previous stage policy checkpoint on a mixture of new-task real demonstrations and \methodabbrev-generated pseudo-trajectories of previous tasks.  We provide the task ordering used for each LIBERO benchmark below, covering both the base-stage tasks and the order in which tasks are introduced during the continual learning stages. The same ordering is used across all compared methods for fair comparison.

\resizebox{\columnwidth}{!}{%
\begin{tcolorbox}[colback=blue!5, colframe=black, title=\texttt{LIBERO-Object Task Order}, fonttitle=\bfseries,label={box:libero_object_order}]
\textit{Base stage:}
\begin{enumerate}
    \setlength\itemsep{1pt}
    \item Pick up the alphabet soup and place it in the basket.
    \item Pick up the cream cheese and place it in the basket.
    \item Pick up the salad dressing and place it in the basket.
    \item Pick up the BBQ sauce and place it in the basket.
    \item Pick up the ketchup and place it in the basket.
    \item Pick up the tomato sauce and place it in the basket.
\end{enumerate}
\tcblower
\textit{Continual learning stage:}
\begin{enumerate}
    \setlength\itemsep{1pt}
    \setcounter{enumi}{6}
    \item Pick up the butter and place it in the basket.
    \item Pick up the milk and place it in the basket.
    \item Pick up the chocolate pudding and place it in the basket.
    \item Pick up the orange juice and place it in the basket.
\end{enumerate}
\end{tcolorbox}%
}

\resizebox{\columnwidth}{!}{%
\begin{tcolorbox}[colback=blue!5, colframe=black, title=\texttt{LIBERO-Goal Task Order}, fonttitle=\bfseries, label={box:libero_goal_order}]
\textit{Base stage:}
\begin{enumerate}
    \setlength\itemsep{1pt}
    \item Open the middle drawer of the cabinet.
    \item Put the bowl on the stove.
    \item Put the wine bottle on top of the cabinet.
    \item Open the top drawer and put the bowl inside.
    \item Put the bowl on top of the cabinet.
    \item Push the plate to the front of the stove.
\end{enumerate}
\tcblower
\textit{Continual learning stage:}
\begin{enumerate}
    \setlength\itemsep{1pt}
    \setcounter{enumi}{6}
    \item Put the cream cheese in the bowl.
    \item Turn on the stove.
    \item Put the bowl on the plate.
    \item Put the wine bottle on the rack.
\end{enumerate}
\end{tcolorbox}%
}

\resizebox{\columnwidth}{!}{%
\begin{tcolorbox}[colback=blue!5, colframe=black, title=\texttt{LIBERO-Spatial Task Order} , fonttitle=\bfseries , label={box:libero_spatial_order}]
\textit{Base stage:}
\begin{enumerate}
    \setlength\itemsep{1pt}
    \item Pick up the black bowl between the plate and the ramekin and place it on the plate.
    \item Pick up the black bowl next to the ramekin and place it on the plate.
    \item Pick up the black bowl from table center and place it on the plate.
    \item Pick up the black bowl on the cookie box and place it on the plate.
    \item Pick up the black bowl in the top drawer of the wooden cabinet and place it on the plate.
    \item Pick up the black bowl on the ramekin and place it on the plate.
\end{enumerate}
\tcblower
\textit{Continual learning stage:}
\begin{enumerate}
    \setlength\itemsep{1pt}
    \setcounter{enumi}{6}
    \item Pick up the black bowl next to the cookie box and place it on the plate.
    \item Pick up the black bowl on the stove and place it on the plate.
    \item Pick up the black bowl next to the plate and place it on the plate.
    \item Pick up the black bowl on the wooden cabinet and place it on the plate.
\end{enumerate}
\end{tcolorbox}%
}

\subsection{Training Hyperparameters}

Table~\ref{tab:hyperparameters} presents the detailed hyperparamters used during training and inference in all our simulation and real-world experiments. 

\begin{table}[h!]
\centering
\caption{Hyperparameters used in \methodabbrev framework.}
\label{tab:hyperparameters}
\small
\begin{tabular}{@{}lc@{}}
\toprule
\textbf{Hyperparameter} & \textbf{Value} \\
\midrule
\rowcolor{gray!50}\multicolumn{2}{l}{\emph{\textbf{Training}}} \\
Input image resolution & $224 \times  224 $ \\
Action dim  & 7 \\
Proprio dim & 9 \\
Action chunk ($H$) & $16$ \\
Use wrist image & True \\
Normalize actions, proprio & True \\
Optimizer & Adam \\
Peak learning rate & $1 \times 10^{-4}$ \\
Warm-up steps & $1{,}000$ \\
Cosine decay steps & $30{,}000$ \\
Batch size per GPU & $40$  \\
No. of GPUs & $4$ \\ 
Gradient accumulation steps & $12$ \\
Augmentations & random crop, color jitter, gaussian blur \\
Base-stage iterations & $10K$ \\
CL-stage iterations (per stage) & $2K$ \\
Max replay episodes per previous task & $10$ \\
\midrule
\rowcolor{gray!50}\multicolumn{2}{l}{\emph{\textbf{Inference}}} \\

Denoising steps (actions) & $5$ \\
Denoising steps (observations, value) & $1$ \\
Action chunk & $16$ \\
\bottomrule
\end{tabular}
\end{table}

\subsection{Evaluation}

\textbf{LIBERO.} After each continual learning stage, we evaluate the policy on all tasks observed up to that point. For each task, we run $50$ trials with randomized initial states  and report the average success rate. From these per-task success rates, we compute the three continual learning metrics: FWT, NBT, and AUC as defined in Sec.~\ref{sec:implementation}.

\textbf{Real-world.} We evaluate on three  real-world manipulation tasks introduced sequentially, with $10$ trials per task from randomized object placements and initial gripper configurations. Rollouts are scored using a partial-scoring rubric: $50$ points for touching the target object and $50$ points for reaching the goal.

\section{Additional Visualizations}
\subsection{Examples of generated pseudo-trajectories}
\label{app:qualitative-examples}
\begin{figure}[h!]
    \centering
    \begin{subfigure}{\linewidth}
        \centering
        \includegraphics[width=\linewidth]{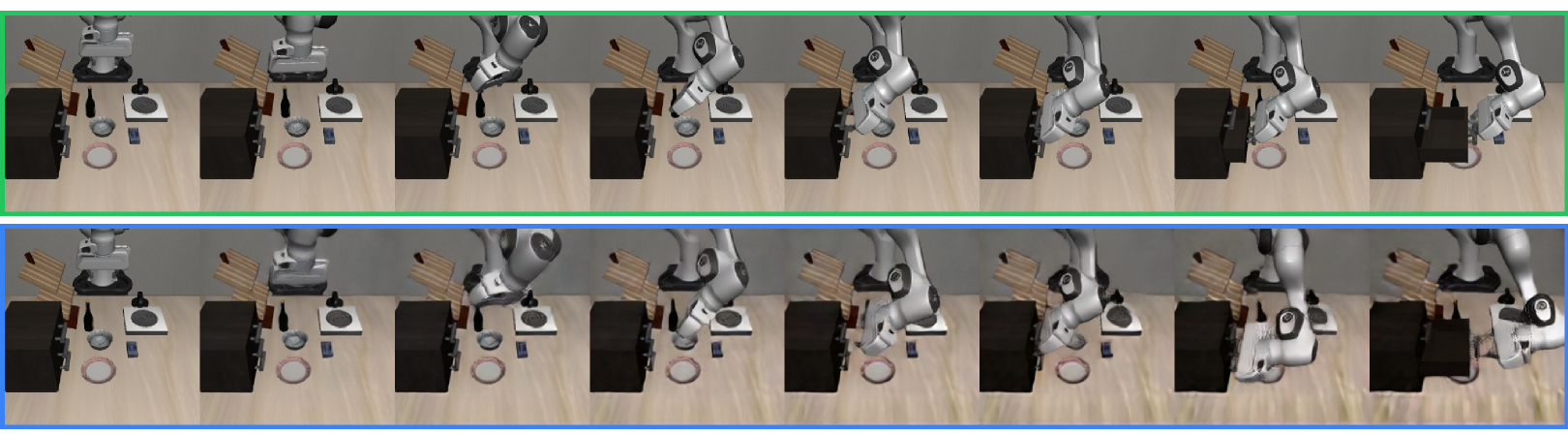}
        \caption{\emph{Open the middle drawer of the cabinet.}}
        \label{fig:qual_goal_2}
    \end{subfigure}
    \label{fig:qual_goal}
     \vspace{-1em}
    \begin{subfigure}{\linewidth}
        \centering
        \includegraphics[width=\linewidth]{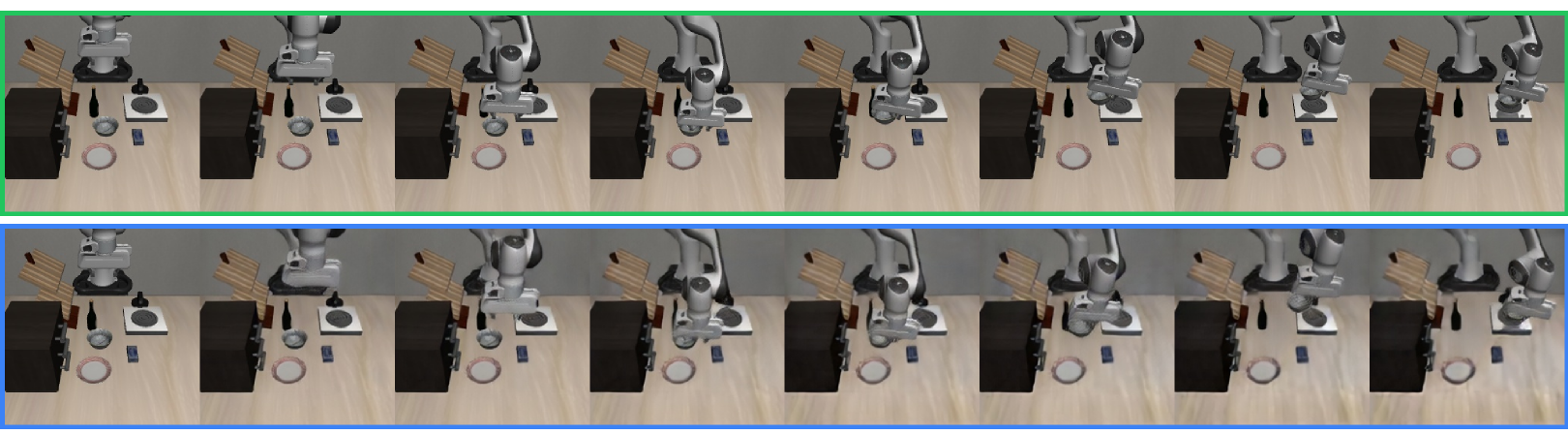}
        \caption{\emph{Put the bowl on the stove.}}
        \label{fig:qual_goal_2}
    \end{subfigure}
    \label{fig:qual_goal}
     \vspace{-1em}
    \begin{subfigure}{\linewidth}
        \centering
        \includegraphics[width=\linewidth]{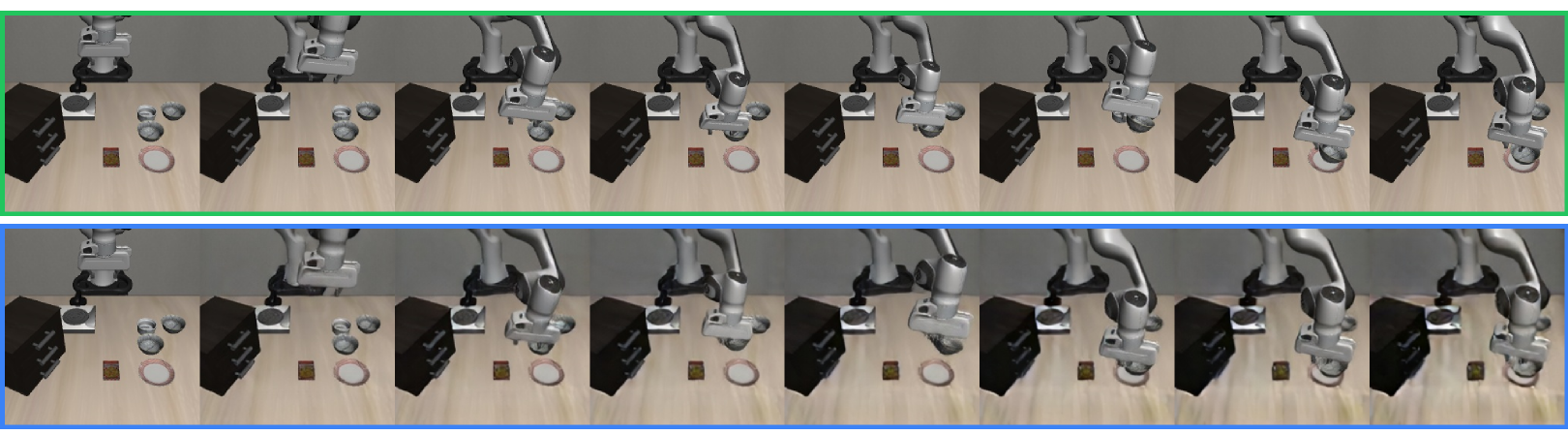}
       \caption{\emph{Pick up the black bowl between the plate and the ramekin and place it on the plate.}}
        \label{fig:qual_goal_2}
    \end{subfigure}
     \vspace{-0.5em}
    \begin{subfigure}{\linewidth}
        \centering
        \includegraphics[width=\linewidth]{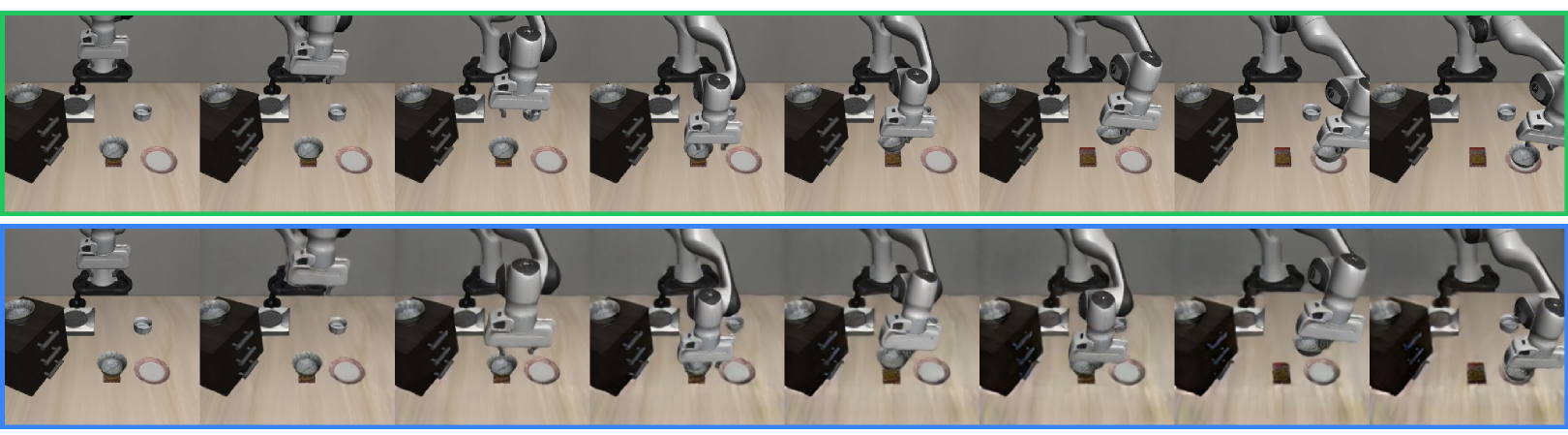}
       \caption{\emph{Pick up the black bowl on the cookie box and place it on the plate.}}
        \label{fig:qual_goal_2}
    \end{subfigure}
    \caption{\textbf{Visualization of pseudo-trajectories.} \textbf{\textcolor[HTML]{22c55e}{Green}} frames represents ground-truth expert demonstrations and \textbf{\textcolor[HTML]{3b82f6}{blue}} frames correspond to pseudo-trajectories generated by \methodabbrev.}
    \label{fig:regen_goal_examples}
\end{figure}

In Figure~\ref{fig:regen_goal_examples}, we visualize pseudo-trajectories generated by \methodabbrev are visually similar to true expert demonstrations, supporting the use of these synthesized trajectories as replay data.

\subsection{Qualitative results of~\methodabbrev}
\label{app:qualitative-results}

\begin{figure}[h!]
    \centering
    \begin{subfigure}{\linewidth}
        \centering
        \includegraphics[width=0.85\linewidth]{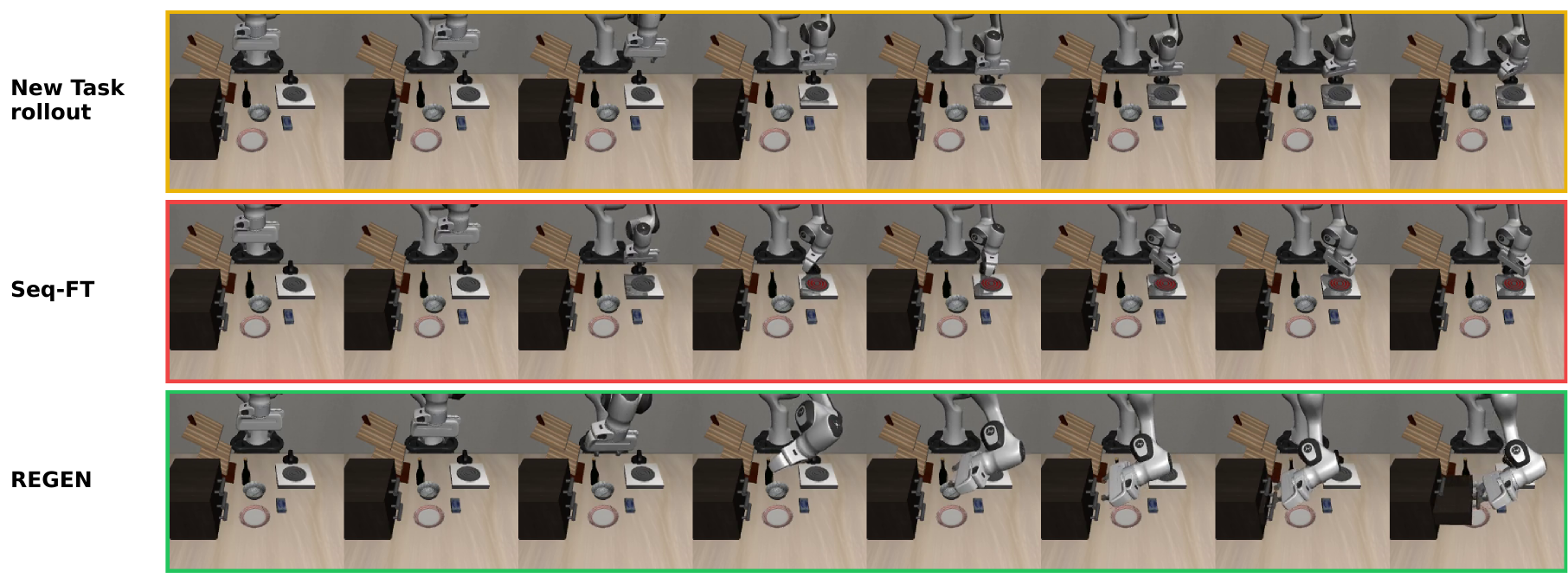}
       \caption{\emph{LIBERO-Goal: evaluation on task 1 (``open the middle drawer of the cabinet'') after training on task 8 (``turn on the stove'').}}
        \label{fig:qual_goal}
    \end{subfigure}

    \begin{subfigure}{\linewidth}
        \centering
        \includegraphics[width=0.85\linewidth]{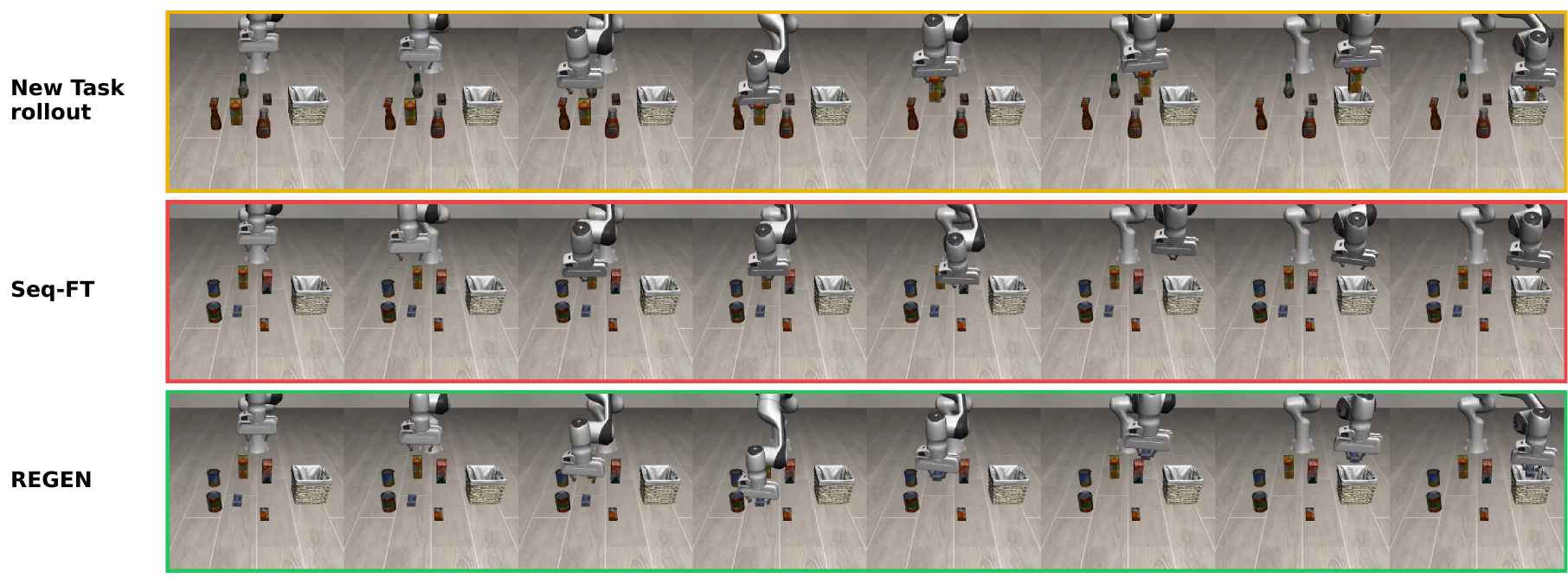}
        \caption{\emph{LIBERO-Object: evaluation on task 2 (``Pick up the cream cheese and place it in the basket.'') after training on task 10 (``pick up the orange juice and place it in the basket'').}}
        \label{fig:qual_object}
    \end{subfigure}
    \setcounter{subfigure}{2}
    \begin{subfigure}{\linewidth}
        \centering
        \includegraphics[width=0.85\linewidth]{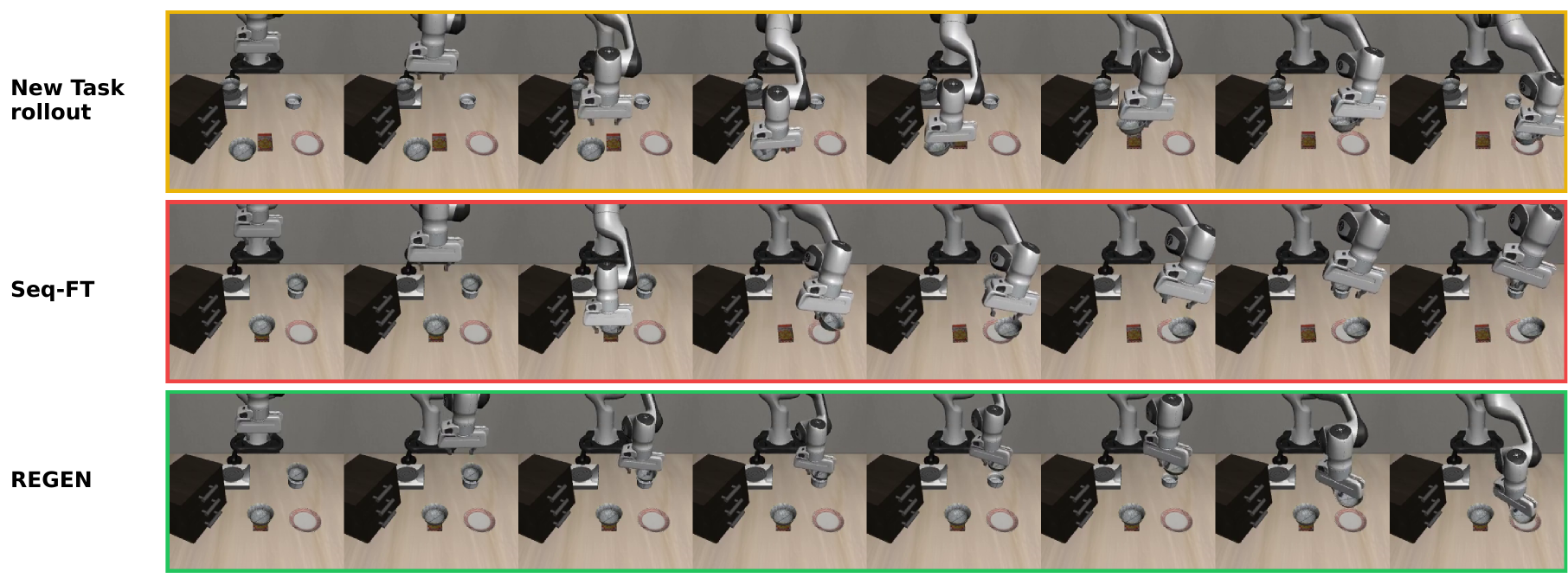}
        \caption{\emph{LIBERO-Spatial: evaluation on task 6 (``Pick up the black bowl on the ramekin and place it on the plate.'') after training on task 7 (``Pick up the black bowl next to the cookie box and place it on the plate.'').}}
        \label{fig:qual_spatial}
    \end{subfigure}

    \begin{subfigure}{\linewidth}
        \centering
        \includegraphics[width=0.85\linewidth]{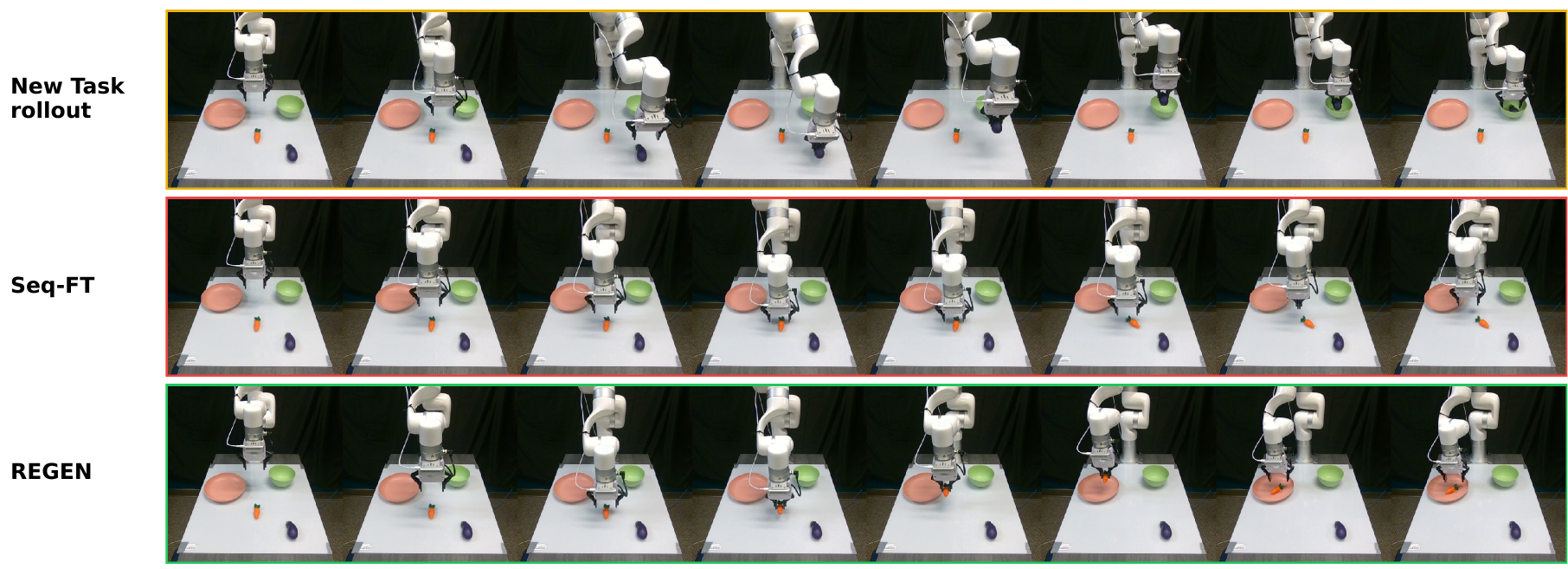}
        \caption{\emph{Real-world: evaluation on Task 2 (``put the carrot on the plate'') after training on Task 3 (``put the eggplant in the bowl'').}}
        \label{fig:qual_realworld}
    \end{subfigure}
    \caption{\textbf{Qualitative comparison on previously seen tasks after continual learning.} In (a)-(d), \textbf{\textcolor[HTML]{eab308}{Top:}}current CL-stage task rollout. \textbf{\textcolor[HTML]{ef4444}{Middle:}} Seq-FT rollout on the previous task, demonstrating catastrophic forgetting by executing the current task instead or failing to accomplish the previous task. \textbf{\textcolor[HTML]{22c55e}{Bottom:}} \methodabbrev successful rollouts on the previous task, retaining task-relevant behavior.}
    \label{fig:qual_libero_real_world_results}
\end{figure}

Figure~\ref{fig:qual_libero_real_world_results} compares qualitative rollouts of \methodabbrev and Seq-FT on previous tasks across LIBERO and the real-world benchmarks.

\subsection{Visualization of \methodabbrev trajectories across continual learning stages}
\label{app:obs_quality}

\begin{figure}[ht!]
    \centering
    \includegraphics[width=\linewidth]{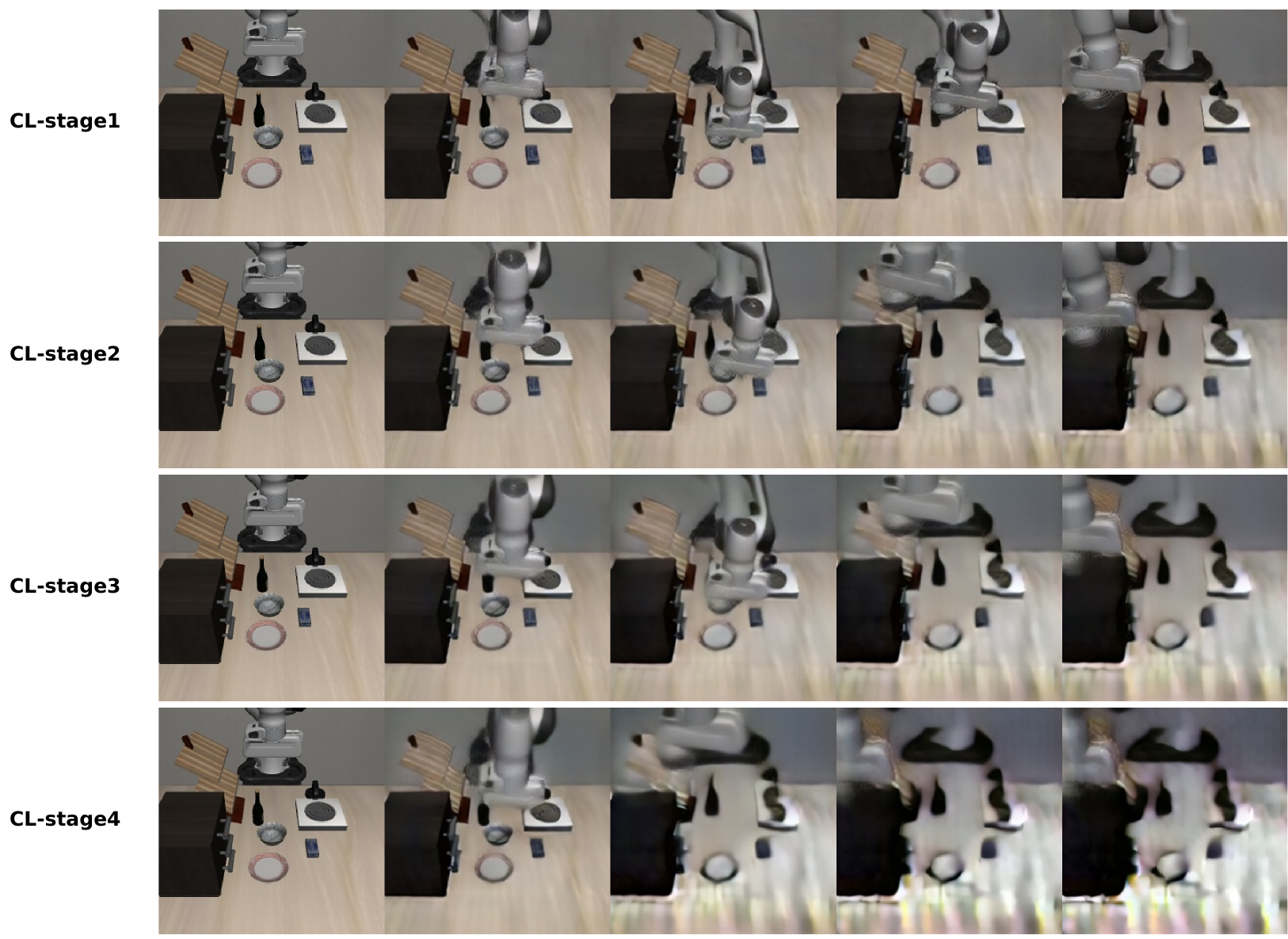}
\caption{\textbf{Degradation of pseudo-trajectory visual observation  quality across CL stages.} Generated trajectories for the task \emph{``put the bowl on top of the cabinet''} show progressively increasing blur.}
\label{fig:obs_quality}
\end{figure}

We visualize the pseudo-trajectories generated by \methodabbrev across successive continual learning stages in Figure~\ref{fig:obs_quality}.
\subsection{Inconsistency between predicted observations and actions}
\label{app:action_obs_mismatch}
\begin{figure}[h!]
    \centering
    \begin{subfigure}{\linewidth}
        \centering
        \includegraphics[width=\linewidth]{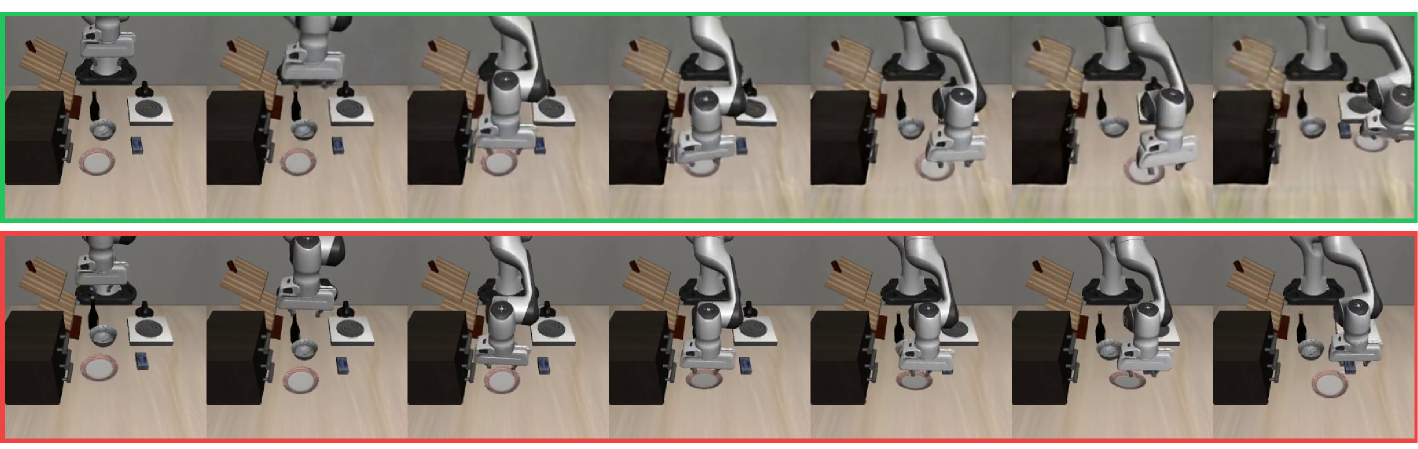}
        \caption{\emph{Push the plate infront of the stove}}
    \end{subfigure}
    \vspace{1pt}
    \begin{subfigure}{\linewidth}
        \centering
        \includegraphics[width=\linewidth]{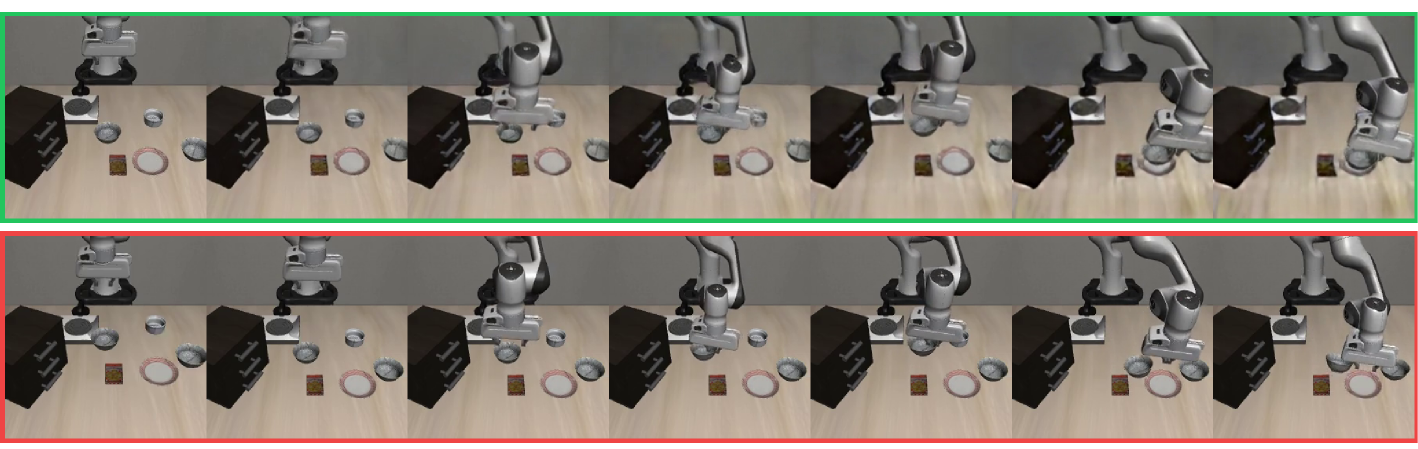}
        \caption{\emph{Pick up the black bowl from table center and place it on the plate.}}
    \end{subfigure}
    \caption{\textbf{Inconsistency between predicted observations and actions.} \textbf{\textcolor[HTML]{22c55e}{Top row:}} future observations imagined by the WAM, which appear to successfully complete the task. \textbf{\textcolor[HTML]{ef4444}{Bottom row:}} executing the predicted actions in the simulator fails to represent the imagined future observation, revealing that the WAM generates visually plausible outcomes without ensuring that the corresponding actions are physically sufficient.}
    \label{fig:action_obs_mismatch}
\end{figure}

We show representative examples of the inconsistency between the WAM's predicted observations and actions in Figure~\ref{fig:action_obs_mismatch}.

\end{document}